\documentclass[conference]{IEEEtran}
\IEEEoverridecommandlockouts

\usepackage{xparse}
\usepackage[utf8]{inputenc} 
\usepackage[T1]{fontenc}    
\usepackage{hyperref}       
\usepackage{url}            
\usepackage{booktabs}       
\usepackage{nicefrac}       
\usepackage{xcolor}         
\usepackage{bm}
\usepackage[noadjust]{cite}
\usepackage{graphicx}
\graphicspath{{figures/}}
\usepackage{hyperref}
\hypersetup{colorlinks,linkcolor={blue},citecolor={green},urlcolor={blue}}  
\usepackage{caption}
\usepackage{multirow}
\usepackage[flushleft]{threeparttable}
\usepackage{comment}
\usepackage{amsmath}
\usepackage{amssymb}
\usepackage{mathtools}
\usepackage[font=small,skip=5pt]{caption}
\usepackage[font=small,position=b,skip=5pt]{subcaption}
\usepackage[capitalise]{cleveref}
\usepackage{leftidx}
\usepackage[normalem]{ulem}

\NewDocumentCommand\Vector{m}{ \boldsymbol{\mathbf{#1}} }
\NewDocumentCommand\Matrix{m}{ \boldsymbol{\mathbf{#1}} }
\NewDocumentCommand\ArgMin{m}{ \operatorname*{min}_{#1} }
\NewDocumentCommand\Real{}{ \mathbb{R} }
\NewDocumentCommand\LieGroupSO{m}{ \mathrm{SO}(#1) }

\NewDocumentCommand\LieGroupSE{m}{ \mathrm{SE}(#1) }

\NewDocumentCommand\ZeroMatrix{}{ \Matrix{0} }
\NewDocumentCommand\IdentityMatrix{}{ \Matrix{1} }
\NewDocumentCommand\CoordinateFrame{m}{ \underrightarrow{\Matrix{\mathcal{F}}}_{#1} }

\NewDocumentCommand\Matlog{m}{\mathrm{log}\left({#1}\right)^{\vee}}
\NewDocumentCommand\Matexp{m}{\mathrm{exp}\left({#1}^{\wedge}\right)}

\NewDocumentCommand\LeftJacobianSO{}{ \Matrix{J}_\ell }

\NewDocumentCommand\CCal{}{{\mathbf{C}_{rc}}}
\NewDocumentCommand\CCalT{}{{\mathbf{C}^\top_{rc}}}
\NewDocumentCommand\rCal{}{{\mathbf{r}^{cr}_{r}}}

\NewDocumentCommand\rMeas{}{\tilde{\boldsymbol{r}}^{c_{k+1}c_k}_{c_k}}
\NewDocumentCommand\bbm{}{ \begin{bmatrix} }
\NewDocumentCommand\ebm{}{ \end{bmatrix} }

\title{\LARGE \bf 
A Self-Supervised, Differentiable Kalman Filter for\\
Uncertainty-Aware Visual-Inertial Odometry}

\author{Brandon Wagstaff$^{1}$, Emmett Wise$^{1}$, and Jonathan Kelly$^{1\dagger}$
\thanks{$^1$All authors are with the Space \& Terrestrial Autonomous Robotic Systems (STARS) Laboratory at the University of Toronto Institute for Aerospace Studies (UTIAS), Toronto, Ontario, Canada, M3H~5T6. Email: \texttt{<first name>.<last name>@robotics.utias.utoronto.ca}}
\thanks{$^\dagger$Jonathan Kelly is a Vector Institute Faculty Affiliate. This research was supported in part by the Canada Research Chairs program.}}

\begin{document}
\maketitle
\thispagestyle{empty}
\pagestyle{empty}

\begin{abstract}
Visual-inertial odometry (VIO) systems traditionally rely on filtering or optimization-based techniques for egomotion estimation.
While these methods are accurate under nominal conditions, they are prone to failure during severe illumination changes, rapid camera motions, or on low-texture image sequences.
Learning-based systems have the potential to outperform classical implementations in challenging environments, but, currently, do not perform as well as classical methods in nominal settings.
Herein, we introduce a framework for training a hybrid VIO system that leverages the advantages of learning \emph{and} standard filtering-based state estimation.
Our approach is built upon a differentiable Kalman filter, with an IMU-driven process model and a robust, neural network-derived relative pose measurement model.
The use of the Kalman filter framework enables the principled treatment of uncertainty at training time and at test time.
We show that our self-supervised loss formulation outperforms a similar, supervised method, while also enabling online retraining.
We evaluate our system on a visually degraded version of the EuRoC dataset and find that our estimator operates without a significant reduction in accuracy in cases where classical estimators consistently diverge.  
Finally, by properly utilizing the metric information contained in the IMU measurements, our system is able to recover metric scene scale, while other self-supervised monocular VIO approaches cannot.
\end{abstract}

\section{Introduction}

Maintaining an accurate estimate of a robot's egomotion (i.e., position, velocity, and orientation) in challenging environments remains a difficult problem.
A common method of egomotion estimation is visual-inertial odometry (VIO), which fuses camera and inertial measurement unit (IMU) data to determine the metric pose change between sequential camera frames \cite{gui2015review}.
Traditionally, VIO approaches rely on classical filtering or optimization-based back-ends to estimate the pose change that best explains the relative motion of tracked visual features.
However, these classical VIO feature trackers are often ``brittle'' in challenging scenes and under rapid motion (due to, e.g., motion blur).
Recent work has explored data-driven replacements for classical estimators.

\begin{figure}[t!]
	\centering
	\includegraphics[width=\columnwidth]{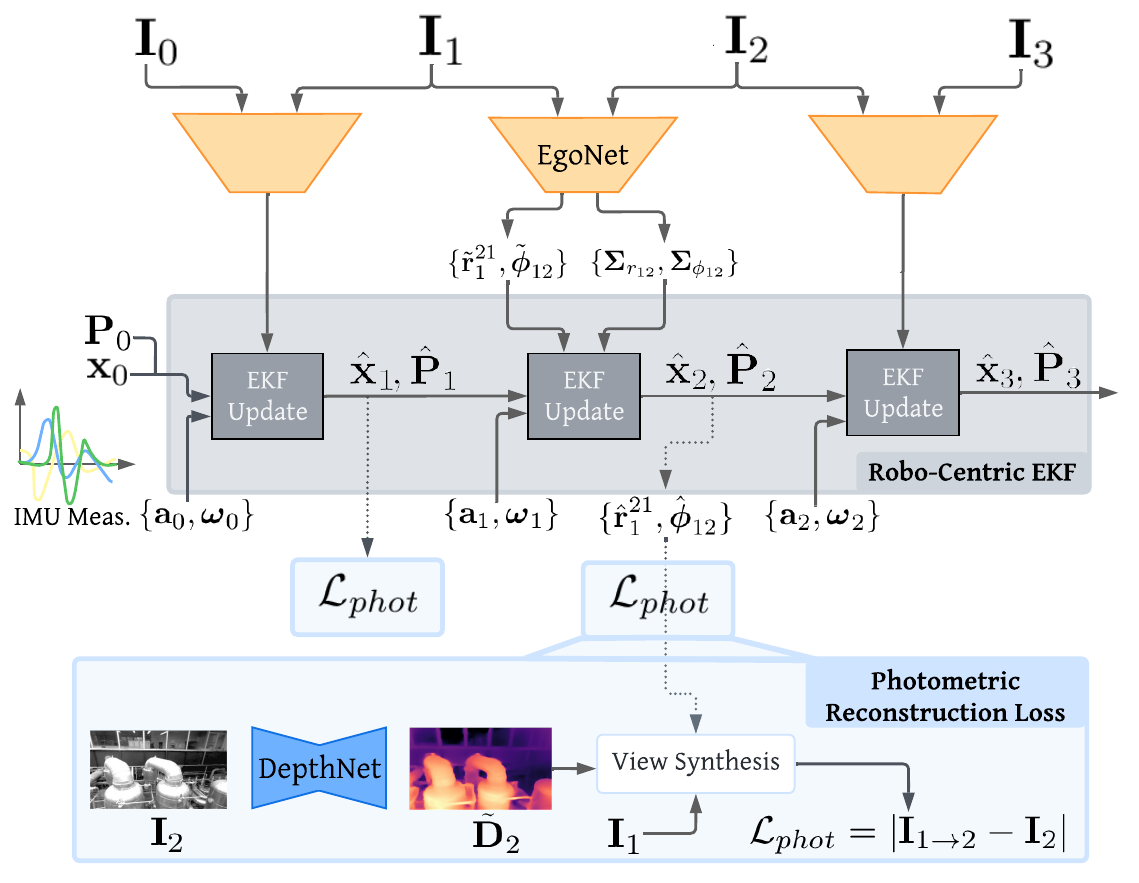}
	\caption{Overview of our hybrid method that combines an IMU-based process model with a learned relative pose measurement model through a robocentric extended Kalman filter (EKF). Our self-supervised formulation can train this system end-to-end by minimizing a photometric reconstruction loss. Unlike basic self-supervised VO or VIO systems that directly use the egomotion prediction for view synthesis, we use the \textit{a posteriori} egomotion estimate produced by the EKF, which refines the network output by incorporating inertial measurements. 
	}
	\label{fig:system-diagram}
	\vspace{-6mm}
\end{figure}
In data-driven systems, neural networks are trained via supervised \cite{gurturk2021ytu,chen2019selective,clark2017vinet} or self-supervised \cite{almalioglu2019selfvio,wei2021unsupervised} learning to model the complex relationship between sensor measurements and robot egomotion.
While neural networks are increasing in popularity and accuracy, they still have limitations.
Networks trained using supervised loss functions struggle to generalize beyond the initial training set.
Generalization improves with the inclusion of more data, but labelling these data is costly.
Consequently, self-supervised loss formulations, which do not require data labelling, are replacing supervised learning in many areas.
However, despite many advances \cite{almalioglu2019selfvio,wei2021unsupervised}, self-supervised approaches do not typically leverage knowledge from the domain of sensor fusion, so network-based methods provide no notion of uncertainty with their predictions.
Furthermore, for VIO, these approaches may not properly utilize scale information contained within the IMU measurements, which is fundamental in classical sensor fusion.

Herein, we propose a novel \textit{hybrid} system that leverages the benefits---and mitigates the limitations---of classical and learning-based VIO systems.
Our sensor fusion scheme (illustrated in \Cref{fig:system-diagram}) uses a differentiable filter (DF) \cite{kloss2021train} to incorporate network-based egomotion measurements into an uncertainty-aware state estimator. We rely on IMU measurements to propagate the state between camera frames.
Our work builds upon that of Li et al.\ \cite{li2020towards}, who used supervised pose labels to train an end-to-end system.
We relax these training requirements by replacing the supervised loss with a pixel-based reconstruction loss that is fully self-supervised. 
Notably, our approach is the first self-supervised, monocular method that produces metrically scaled predictions. 
These predictions are \textit{uncertainty-aware}: we incorporate a learned heteroscedastic uncertainty model, allowing for principled fusion of the measurements within the filter. 
Our experiments demonstrate the robustness of our algorithm to visually degraded environments, relative to the performance of classical systems.
\section{Related Work}

In this section, we separate existing approaches to VIO into three categories: classical, learned, and hybrid, referring to the amount of (or the lack of) machine learning involved in each case.

\subsection{Classical Approaches to VIO}

Generally, VIO algorithms incorporate a front-end and a back-end.
The front-end detects and tracks features across images.
The back-end estimates the 3D locations of tracked features and the camera trajectory. The formulation can be either loosely-coupled or tightly-coupled.
Loosely coupled methods estimate the system state with each sensor modality independently and then combine the estimates in a final stage.
Conversely, tightly-coupled methods incorporate each modality within a joint estimation framework.
To determine the motion, the reprojection error between predicted and observed feature locations is minimized. 
The reprojection error may be expressed indirectly in terms of pixel coordinates, or directly in terms of a photometric (pixel intensity) loss.

One well-known filter-based VIO algorithm is the multi-state constraint Kalman Filter (MSCKF) \cite{mourikis2007multi}, but its world-centric formulation can be inconsistent \cite{li2013high}.
Robot-centric formulations such as ROVIO \cite{bloesch2017iterated} and R-VIO \cite{huai2018robocentric} largely avoid inconsistency.
Optimization-based algorithms such as VINS-Mono \cite{qin2018vins} and OK-VIS \cite{leutenegger2015keyframe} optimize over a window of poses, while relying on IMU preintegration \cite{forster2015preint} for computational efficiency.
Although more compute-intensive, optimization methods tend to be more accurate than their filter-based counterparts.
Herein, we demonstrate how several classical methods (namely VINS-Mono, ROVIO, and R-VIO) are prone to failure in visually degraded scenes.
\subsection{Learning-Based Approaches to VIO}

Recent neural network-based visual-inertial fusion schemes  \cite{gurturk2021ytu,chen2019selective,clark2017vinet,almalioglu2019selfvio,wei2021unsupervised} all adopt a similar ``feature fusion'' procedure, where the neural network learns to map the raw (visual and inertial) measurements to a 6-DOF egomotion prediction in an end-to-end manner.
Internally, the network extracts and combines sensor-specific features (e.g., through concatenation of the two feature vectors).
The resulting multimodal features are then fed into a final network block that  predicts the egomotion.
This scheme exists in both supervised \cite{gurturk2021ytu,chen2019selective,clark2017vinet} and self-supervised \cite{almalioglu2019selfvio,wei2021unsupervised} settings.
In the self-supervised setting \cite{zhou:2017}, a pixel-based reconstruction loss is minimized to jointly train a depth and an egomotion network.
The training process leverages the depth and egomotion predictions to project pixels from a nearby source image into a target frame, in a process known as view synthesis.
The pixel-based reconstruction loss is the per-pixel photometric difference between the projected image and an image taken in the target frame.
Since view synthesis generates a more realistic projected image as the depth and egomotion estimates improve, the networks can be trained by minimizing the pixel-based loss.

While end-to-end approaches are effective, they do not utilize the canonical relationship between inertial measurements and robot dynamics.
Ignoring this relationship burdens the network with learning well-modelled kinematics from scratch and prevents the network from utilizing the metric information in the inertial data.
For example, to utilize the IMU linear acceleration effectively, the network must learn implicitly to track the robot's velocity.
To remove the gravity component from the acceleration measurements, the network must also learn to track the global IMU orientation.
Finally, metric information available from inertial measurements cannot easily be utilized because reconstruction-based losses do not account for absolute scale.
Consequently, the depth and egomotion predictions are only accurate up to a scale factor.
Our aim is to resolve these issues with our hybrid approach that, in contrast to the feature-fusion approach, combines visual and inertial information in a probabilistic manner.
\subsection{Hybrid Approaches to VIO}
Differentiable filters (DFs) have been proposed as a way to impose prior knowledge on  the network structure by combining \textit{perception} (i.e., through a measurement model that maps sensory observations to the state) and \textit{prediction} (i.e., through a process model that determines how the state changes over time) in a Bayesian manner \cite{Haarnoja:2016,kloss2021train}.
Being fully differentiable, DF's can be trained end-to-end to produce uncertainty-aware measurement and process models that accurately account for sensor noise characteristics.
The DF is particularly useful for replacing (brittle) hand-crafted models with networks that directly map high dimensional, nonlinear measurements (e.g., raw images) onto the state.
Applications of this hybrid structure include camera relocalization \cite{zhou2020kfnet}, object tracking \cite{Haarnoja:2016}, and VIO. 
Chen et al.\ \cite{chen2021fully} use a DF to learn the noise parameters of an EKF-based VIO system.
The authors of \cite{chen2021dynanet} train a visual-inertial measurement model for a Kalman filter through a feature fusion-scheme with a learned process model. 
Li et al.\ \cite{li2020towards} present a DF that combines a classical IMU-based process model with a learned relative pose measurement model.
Their network is trained end-to-end by minimizing a pose supervision loss.
We extend the approach of Li et al.\ \cite{li2020towards} by training our network with a photometric reconstruction loss, which improves the overall system accuracy by leveraging the data efficiency of self-supervised learning.
To the best of the authors' knowledge, we are the first to train a differentiable filter in a fully self-supervised manner.
\section{Approach}
\label{sec:approach}

Our hybrid approach to VIO augments the learned depth and egomotion estimation system with a robocentric EKF back-end.
By doing so, the photometric reconstruction loss can be computed with the refined, \textit{a posteriori} egomotion estimate, rather than the direct network output.
This posterior estimate, notably, is a function of the inertial measurements used to propagate the state via the process model.
Since the filter structure is fully differentiable, the whole system can be trained end-to-end by minimizing the photometric reconstruction loss.

In \Cref{sec:notation}, we review the notation used throughout this work.
In Section \ref{sec:depth_and_ego}, we discuss our self-supervised depth and egomotion estimation formulation.
In \Cref{sec:robocentric-ekf}, we review the robocentric EKF for VIO.
Finally, in \Cref{sec:e2e-train}, we present our end-to-end training scheme that minimizes the self-supervised reconstruction loss.
\subsection{Notation} 
\label{sec:notation}

We begin by defining four reference frames that are used throughout \Cref{sec:approach}. 
Let $\CoordinateFrame{i}$, $\CoordinateFrame{c_k}$, $\CoordinateFrame{r_k}$, $\CoordinateFrame{v_\tau}$ represent a static (global) reference frame, the camera reference frame at time $k$, the robocentric reference frame at time $k$, and the IMU reference frame at time $\tau$, respectively.
The scalars $k$ and $\tau$ are the image and IMU time steps (which generally differ).
Ordinary lowercase letters are reserved for scalar quantities.
Bolded lowercase Roman and Greek letters represent vector quantities.
The symbol $\delta$ is used to denote a perturbation to the subsequent quantity. 
The vector $\Vector{g}_{a} \in \Real^3$ is the gravity vector expressed in $\CoordinateFrame{a}$.
The vectors $\Vector{b}_{\omega_k}, \Vector{b}_{a_k}\! \in \Real^3$ are, respectively, the IMU gyroscope and accelerometer biases at time step $k$.
The vectors $\Vector{r}_{a}^{ba}, \Vector{v}_a^{ba}, \Vector{\omega}_{a}^{ba}, \Vector{a}_{a}^{ba}  \in \Real^3$ are the translation, linear velocity, rotational velocity, and linear acceleration of $\CoordinateFrame{b}$ with respect to $\CoordinateFrame{a}$ expressed in $\CoordinateFrame{a}$.
A (normally distributed) noise vector is represented using $\Vector{n}_j \sim \mathcal{N}\left(\Vector{0}, \sigma_j^2\IdentityMatrix\right)$, where $j$ is an appropriate, context-specific identifier.
Bolded uppercase Roman and Greek letters represent matrices. 
We reserve $\Matrix{C}_{ab} \in \LieGroupSO{3}$ for the matrix that rotates vectors from $\CoordinateFrame{b}$ to $\CoordinateFrame{a}$.
The (known) extrinsic transform between $\CoordinateFrame{c}$ and $\CoordinateFrame{r}$, which is constant across timesteps, consists of the rotation matrix $\CCal$ and the translation $\rCal$.
The notation $\Vector{\phi}_{ab} \in \Real^3$ is used to denote the Lie algebra vector corresponding to $\Matrix{C}_{ab}$.
Perturbations of the rotation state are given by $\Matrix{C}_{ab} = \bar{\Matrix{C}}_{ab}\exp(\delta \Vector{\phi}_{ab}^{\wedge})$, where $\bar{(\cdot)}$ is the nominal rotation matrix.
The operator $(\cdot)^{\wedge}$ is the skew-symmetric operator, and the notation $\tilde{(\cdot)}$ denotes an output (or prediction) from one of our networks.

\subsection{Self-Supervised Depth and Egomotion Estimation}
\label{sec:depth_and_ego}

We employ a depth network to produce a depth prediction $\tilde{\mathbf{D}}_t$ for a target image $\mathbf{I}_t$, and an egomotion network to produce an egomotion prediction $\tilde{\mathbf{T}}_{st} \in \LieGroupSE{3} = \{\tilde{\mathbf{r}}^{c_tc_s}_{c_s}, \tilde{\mathbf{C}}_{c_sc_t}\! \in \LieGroupSO{3}\}$ that is an estimate of the 6-DOF pose change between $\CoordinateFrame{c}$ at times $s$ and $t$. Using these predicted quantities, and the known camera intrinsic matrix $\mathbf{K}$, we can generate (through view synthesis) an image $\Matrix{I}_{s\rightarrow t}$ that ``reconstructs'' the target image using the source image pixels. Specifically, each pixel coordinate $\mathbf{u}'$ within $\mathbf{I}_{s\rightarrow t}$ is populated with the pixel intensity of its corresponding location, $\mathbf{u}$, in the nearby source image $\mathbf{I}_s$ using a pinhole camera model $\pi$:
\begin{align}
\label{eq:loss_terms}
\mathbf{I}_{s\rightarrow t}(\mathbf{u}') =\mathbf{I}_s(\mathbf{u}), & \quad \mathbf{u} = \pi(\tilde{\mathbf{T}}_{st}\pi^{-1}(\mathbf{u}')).
\end{align}
\noindent The pinhole projection model maps a 3D point $\mathbf{p}=\displaystyle\begin{bmatrix}x & y & z\end{bmatrix}^\top$ to its pixel coordinate $\mathbf{u}$ through $\pi(\mathbf{p}) = \mathbf{K}\frac{1}{z}\mathbf{p}$. Note that in practice, the reconstructed image $\mathbf{I}_{s\rightarrow t}$ is produced using a spatial transformer \cite{Jaderberg:2015}.

After view synthesis, the photometric reconstruction loss can be computed by comparing the reconstructed image with the known target image  through the standard combination of the $L_1$ loss and the structural similarity (SSIM) loss \cite{wang2004image}:
\begin{equation}
\label{eq:reconstruction_loss}
\mathcal{L}_{phot}(\mathbf{I}_{s\rightarrow t}, \mathbf{I}_t) = (1-\alpha)\left| \mathbf{I}_{s\rightarrow t} - \mathbf{I}_{t}\right| + \alpha \mathcal{L}_\text{SSIM}(\mathbf{I}_{s\rightarrow t}, \mathbf{I}_{t}).
\end{equation}
We use the minimum reconstruction formulation from Godard et al.\ \cite{godard2019digging}, which, for a given target image, builds $\mathcal{L}_{phot}$ with the minimum per-pixel error values from two adjacent source images: 
\begin{align} \label{eq:min_rec_loss}
\vspace{-0.1cm}
\ArgMin{s \in \{t-1,t+1\}} \mathcal{L}_{phot}(\mathbf{I}_{s\rightarrow t}, \mathbf{I}_t).
\vspace{-0.1cm}
\end{align}
The depth and egomotion network weights are jointly trained to minimize this reconstruction loss through gradient descent. In addition to $\mathcal{L}_{phot}$, we employ two auxiliary losses: a depth smoothness loss \cite{godard2017unsupervised} $\mathcal{L}_{smooth}$ and a geometric consistency loss $\mathcal{L}_{consist}$ \cite{bian2019unsupervised}. 
The automasking method from \cite{godard2019digging} and the self-discovered mask from \cite{bian2019unsupervised} are applied to remove unreliable pixels.

We train the tightly-coupled networks from Wagstaff et al.\ \cite{wagstaff2021self}, as they have been shown to significantly boost egomotion accuracy on challenging indoor datasets. Notably, the egomotion network relies on multiple forward passes to iteratively refine the initial egomotion prediction (see \cite{wagstaff2021self} for further details). In \Cref{sec:e2e-train}, we discuss how this network structure is extended to produce uncertainty-aware predictions in order to incorporate these measurements into the EKF. 
\subsection{Robocentric EKF Formulation} \label{sec:robocentric-ekf} 
Our robocentric EKF formulation is based on the approach from Li et al.\  \cite{li2020thesis}. 
An in-depth explanation of the formulation can be found in \cite{li2020thesis,huai2018robocentric}.
In this formulation, the robocentric state, 
represented by $\Vector{x}$, has the following components (in set notation to accommodate the rotation matrices) at the latest IMU measurement time step, $\tau$,
\begin{equation} 
\label{eq:state-def}
\begin{split}
\Vector{x}_\tau = \; & \left\{\begin{matrix}
\Matrix{C}_{r_ki}, &\! \Vector{r}_{r_k}^{i r_k}, &\! \Vector{g}_{r_k} &\! \mid
\end{matrix}\right. \\
& \mspace{9mu} \left.\begin{matrix}
\Matrix{C}_{r_kv_\tau}, &\! \Vector{r}_{r_k}^{v_\tau r_k}, &\! \Vector{v}_{v_\tau}^{v_\tau i}, &\! \Vector{b}_{\omega_\tau}, &\! \Vector{b}_{a_\tau}
\end{matrix}\right\},
\end{split}
\end{equation}
where the vertical bar separates the \textit{robot} states, $\Vector{x}_{r_k i}$, and the \textit{IMU} states, $\Vector{x}_{r_k v_\tau}$.
The error state vector is 
\begin{equation} \label{eq:state-def}
\begin{split}
\delta\Vector{x}_\tau = \; & \left[\begin{matrix}
\delta\Vector{\phi}_{r_ki}^\top & \delta\Vector{r}_{r_k}^{i r_k \top} & \delta\Vector{g}_{r_k}^\top & \mid
\end{matrix}\right. \\
\; & \; \left.\begin{matrix}
\delta\Matrix{\phi}_{r_kv_\tau}^\top & \delta\Vector{r}_{r_k}^{v_\tau r_k \top} & \delta\Vector{v}_{v_\tau}^{v_\tau i \top} & \delta\Vector{b}_{\omega_\tau}^\top & \delta\Vector{b}_{a_\tau}^\top
\end{matrix}\right]^\top.
\end{split}
\end{equation}

\subsubsection{Process Model and Covariance Propagation} 

To determine the error state vector process model, the IMU measurement model and time derivatives of the IMU states are required.
The IMU measurement model is
\begin{equation} \label{eq:imumm}
\begin{split}
\Vector{a}_{m, \tau} = & \Vector{a}_{v_\tau}^{v_\tau i} + \Matrix{C}_{v_\tau r_k}\Vector{g}_{r_k} + \Vector{b}_{a_\tau} + \Vector{n}_a, \\
\Vector{\omega}_{m,\tau} = & \Vector{\omega}_{v_\tau}^{v_\tau i} + \Vector{b}_{\omega_\tau} + \Vector{n}_\omega.
\end{split} 
\end{equation}
The time derivatives of the IMU states (again using set notation) are
\begin{equation}\label{eq:dotstate}
\begin{split}
\dot{\Vector{x}}_{r_k v_\tau}= & \left\{\begin{matrix} \Matrix{C}_{r_kv_\tau}(\Vector{\omega}_{v_\tau}^{v_\tau i })^\wedge, & \Matrix{C}_{r_kv_\tau}\Vector{v}_{v_\tau}^{v_\tau i}\end{matrix},\right. \\
& \mspace{9mu} \left.\begin{matrix}\Vector{a}_{v_\tau}^{v_\tau i} - (\Vector{\omega}_{v_\tau}^{v_\tau i})^\wedge \Vector{v}_{v_\tau}^{v_\tau i}, & \Vector{n}_{b_\omega}, & \Vector{n}_{b_a} \end{matrix}\right\}.
\end{split}
\end{equation}

By applying perturbations to the state, substituting in Equations \ref{eq:imumm} and \ref{eq:dotstate}, and linearizing the result, the time derivative of the error state vector is
\begin{equation}
\delta\dot{\Vector{x}}_\tau = \Matrix{F}\delta\Vector{x}_\tau + \Matrix{G}\Vector{n},
\end{equation} 
where $\Vector{n} = \bbm \Vector{n}_\omega^\top & \Vector{n}_a^\top & \Vector{n}_{b_\omega}^\top & \Vector{n}_{b_a}^\top \ebm^\top$ is the vector of noise terms. The matrices $\Matrix{F}$ and $\Matrix{G}$ can be found in \cite{huai2018robocentric}.

The process model for the IMU states is used within the prediction step to propagate the state estimate $\hat{\Vector{x}}_k$, expressed with respect to the most recent robocentric frame $\CoordinateFrame{r_k}$, forward from time step $k$ to time step $\tau$ using the IMU measurements.
This procedure yields the predicted IMU state, $\check{\Vector{x}}_{r_k v_{\tau}}$, 
\begin{align}
\check{\Matrix{C}}_{r_{k}v_{\tau}} & = \int_{t_{k}}^{t_{\tau}} 
\check{\Matrix{C}}_{r_kv_s}\left(\Vector{\omega}_{m,s} - \hat{\Vector{b}}_{\omega_{k}}\right)^\wedge ds, \\[1mm]
\begin{split}
\check{\Vector{v}}_{v_{\tau}}^{v_{\tau}i} & = 
\check{\Matrix{C}}^{\top}_{r_{k}v_{\tau}}
\left(\hat{\Vector{v}}_{r_k}^{v_k i} - \hat{\Vector{g}}_{r_k}\Delta t \right. \\ 
& \left.\mspace{100mu} + \int_{t_{k}}^{t_{\tau}} 
\check{\Matrix{C}}_{r_kv_s}\left(\Vector{a}_{m,s} - \hat{\Vector{b}}_{a_{k}}\right) ds\right),
\end{split} \\
\begin{split}
\check{\Vector{r}}_{r_k}^{v_{\tau}r_k} & = \hat{\Vector{v}}_{r_k}^{v_k i}\Delta t - \frac{1}{2}\hat{\Vector{g}}_{r_k}\Delta t^2 \\ 
& \mspace{100mu} + \int_{t_k}^{t_\tau} \int_{t_k}^s \check{\Matrix{C}}_{r_kv_\mu}(\Vector{a}_{m,\mu} - \hat{\Vector{b}}_{a_k})\, d\mu ds.
\end{split}
\end{align}
where $\Delta t = t_{\tau} - t_{k}$.
Discrete integration of the process model is performed using Euler's method. To propagate the state covariance forward in time, we require the transition matrix for the error state between IMU time steps,
\begin{equation}
\Matrix{\Phi}_{\tau + 1, \tau} = \exp\left(\int\limits_{t_\tau}^{t_{\tau + 1}} \Matrix{F}(s) \, ds\right) \approx \IdentityMatrix + \Matrix{F}\delta t,
\end{equation} 
where $\IdentityMatrix$ is the identity matrix and $\delta t = t_{\tau + 1} - t_{\tau}$. The predicted state uncertainty is then 
\begin{align}
\Matrix{Q} & = \text{diag}(\sigma_\omega^2 \IdentityMatrix, \sigma_a^2\IdentityMatrix, \sigma_{b_\omega}^2 \IdentityMatrix, \sigma_{b_a}^2\IdentityMatrix), \\[1mm]
\check{\Matrix{P}}_{\tau + 1} & = \Matrix{\Phi}_{\tau + 1, \tau}\check{\Matrix{P}}_{\tau}\Matrix{\Phi}_{\tau + 1, \tau}^\top + \Matrix{G}\Matrix{Q}\Matrix{G}^\top\delta t.
\end{align}

\subsubsection{Measurement Update} Our measurements are the relative pose changes (egomotion) produced by our network.
Unlike current supervised loss formulations, our self-supervised loss requires the egomotion predictions to be camera-centric (i.e., expressed in $\CoordinateFrame{c_k}$).
The measurement residual, $\Vector{\epsilon}_{k+1} = \begin{bmatrix} \Vector{\epsilon}^\top_\phi & \Vector{\epsilon}^\top_r \end{bmatrix}^\top$, with covariance $\Matrix{R}_k$ (and evaluated with the predicted IMU state at $t_{k+1}$) therefore is 
\begin{align} \label{eq:meas_residual}
\begin{bmatrix} \Vector{\epsilon}_\phi \\ \Vector{\epsilon}_r \end{bmatrix} = \begin{bmatrix} 		
\Matlog{\tilde{\Vector{C}}_{c_{k} c_{k+1}} \check{\Vector{C}}^\top_{c_{k}c_{k+1}}}\\[1mm]
\rMeas - \check{\Vector{r}}^{c_{k+1}c_k}_{c_k}
\end{bmatrix},
\end{align}
where
\begin{align}
\check{\Vector{r}}^{c_{k+1}c_k}_{c_k} &= \CCalT \check{\mathbf{C}}_{r_kv_{k+1}} \rCal + \CCalT \left( \check{\Vector{r}}_{r_k}^{v_{k+1} r_k} - \rCal \right), \\
\check{\Vector{C}}_{c_{k}c_{k+1}} &= \CCalT \check{\Vector{C}}_{r_kv_{k+1}} \CCal.
\end{align}
The measurement Jacobian $\mathbf{H}_{k+1}$ is found by differentiating \Cref{eq:meas_residual} with respect to $\delta\check{\Vector{x}}_{k+1}$: 
\begin{align}
\mathbf{H}_{k+1} = \begin{bmatrix} \ZeroMatrix_{3 \times 9} & -\CCalT\,\LeftJacobianSO (-\check{\phi}_{r_kv_{k+1}}) & \ZeroMatrix_{3 \times 3} & \ZeroMatrix_{3 \times 9} \\[1mm]
\ZeroMatrix_{3 \times 9} & \CCalT\,\check{\mathbf{C}}_{r_kv_{k+1}} \rCal^\wedge & -\CCalT & \ZeroMatrix_{3 \times 9}
\end{bmatrix}.
\end{align}
\noindent Note that the derivation for $\mathbf{H}_{k+1}$ uses the Baker-Campbell-Hausdorff (BCH) formula,
\begin{align}
\Matlog{\Matexp{\Vector{\phi}_2} \Matexp{\Vector{\phi}_{1}}^\top} \approx \Vector{\phi}_2 - \Vector{\phi}_{1}, 
\end{align}
which is reasonable for odometry applications.
Using the derived equations, the EKF measurement update is
\begin{align}
\Matrix{K}_{k+1} = & \check{\Matrix{P}}_{k+1}\Matrix{H}^\top_{k+1}\left(\Matrix{H}_{k+1}\check{\Matrix{P}}_{k+1}\Matrix{H}^\top_{k+1} + \Matrix{R}_{k+1}\right)^{-1}, \nonumber \\
\hat{\Matrix{P}}_{k+1} = & \left(\IdentityMatrix - \Matrix{K}_{k+1}\Matrix{H}_{k+1}\right)\check{\Matrix{P}}_{k+1}, \\
\delta \hat{\Vector{x}}_{k+1} = & \Matrix{K}_{k+1}\Vector{\epsilon}_{k+1}. \nonumber
\end{align}
Finally, the \textit{a posteriori} state $\hat{\Vector{x}}_{k+1}$ is produced by injecting the error-state estimate $\delta \hat{\Vector{x}}_{k+1}$ into the nominal state.
\subsubsection{Composition Step} In the robocentric formulation, the robot state is shifted forward from $\CoordinateFrame{r_{k}}$ to $\CoordinateFrame{r_{k+1}}$, following the measurement update, by compounding the (relative) IMU pose with the robot pose, and updating the gravity vector direction. Then, the IMU pose, which is expressed relative to the robot pose, is reset to the identity: 
\begin{align}
&\hat{\Matrix{C}}_{r_{k+1}i} = \hat{\Matrix{C}}_{r_kv_{k+1}}^\top\hat{\Matrix{C}}_{r_ki}, \hspace{0.4cm}
\hat{\Vector{g}}_{r_{k+1}} = \hat{\Matrix{C}}_{r_kv_{k+1}}^\top\hat{\Vector{g}}_{r_k}, \nonumber \\
&\hat{\Vector{r}}_{r_{k+1}}^{ir_{k+1}} =  \hat{\Matrix{C}}_{r_kv_{k+1}}^\top\left( \hat{\Vector{r}}_{r_{k}}^{ir_{k}} - \hat{\Vector{r}}_{r_{k}}^{v_{k+1}r_{k}}\right), \\
&\hat{\Matrix{C}}_{r_{k+1}v_{k+1}} =  \IdentityMatrix, \hspace{0.4cm}
\hat{\Vector{r}}_{r_{k+1}}^{v_{k+1}r_{k+1}} =  \Vector{0}. \nonumber
\end{align}
This process is referred to as the composition step. The state covariance is accordingly updated using
\begin{equation}
\hat{\Matrix{P}}_{k+1} = \Matrix{U}_{k+1}\hat{\Matrix{P}}_{k+1}\Matrix{U}_{k+1}^\top, \; \; \Matrix{U}_{k+1}=\frac{\partial\delta\hat{\Vector{x}}_{k+1,r_{k+1}}}{\partial\delta\hat{\Vector{x}}_{k+1,r_{k}}}.
\end{equation}
The derivation and value of $\Matrix{U}_{k+1}$ can be found in \cite{huai2018robocentric}.

\subsection{End-to-End Training with the Differentiable EKF} 
\label{sec:e2e-train}

\begin{figure}[t!]
	\centering
	\includegraphics[width=\columnwidth]{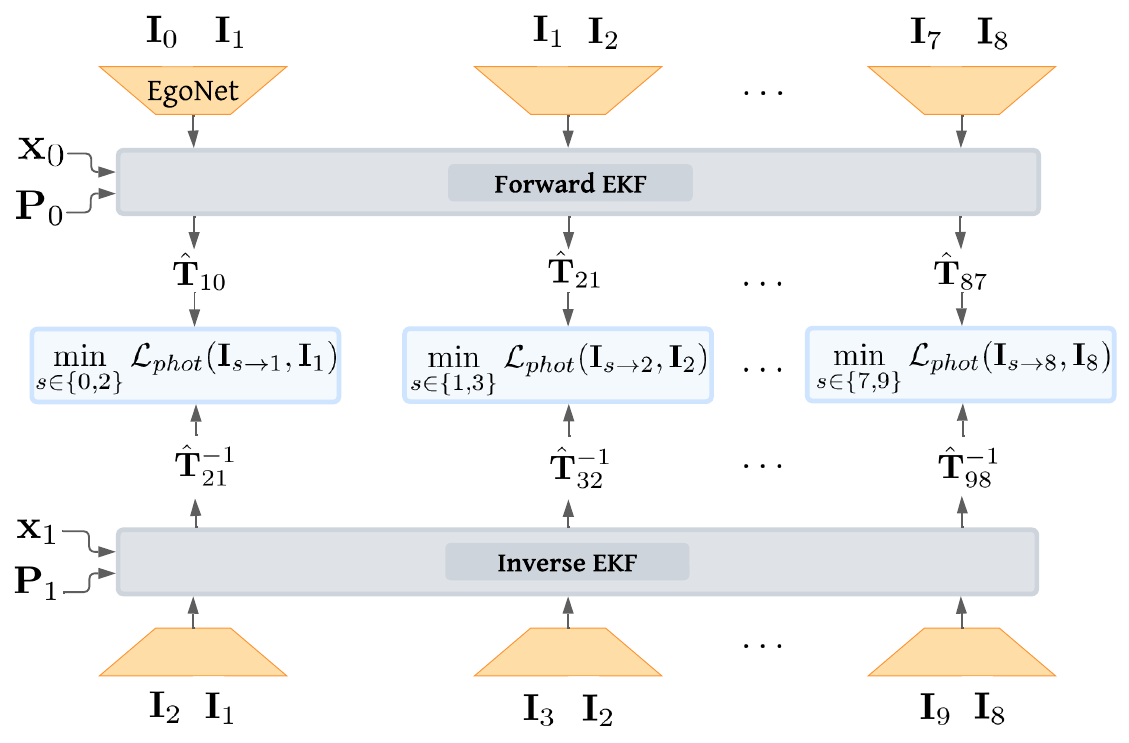}
	\caption{When applying the minimum reconstruction loss from \cite{godard2019digging}, two (adjacent) source images are projected into each target frame. We use the egomotion estimates from a forward and inverse EKF to produce these two reconstructions. Samples with $N$ frames can produce $N-2$  reprojection losses.}
	\label{fig:min_rec_loss}
	\vspace{-5mm}
\end{figure}

\Cref{fig:min_rec_loss} visualizes the training procedure of our hybrid system.
We extend the number of frames per sample from the standard of three (i.e., one target image and two adjacent source images) to an arbitrary length $N$.
The longer sequence length is important for uncertainty learning because the network must learn to inflate the covariance for erroneous measurements that negatively impact future state estimates. 
By reducing the impact of erroneous (e.g., corrupted or outlier) measurements, future estimates will be more accurate and the reconstruction loss will be reduced. 

The EKF propagates from the initial state $\mathbf{x}_0$ and incorporates the learned egomotion measurements.
Then, the \textit{a posteriori} IMU pose is used to compute the minimum reconstruction loss (\Cref{eq:min_rec_loss}).
This loss requires two image reconstructions.
The first uses the ``forward'' egomotion predictions to produce $\mathbf{I}_{t-1\rightarrow t}$.
The second uses the ``inverse'' predictions (i.e., the source-target image inputs are swapped) to produce $\mathbf{I}_{t+1\rightarrow t}$.
As demonstrated in \Cref{fig:min_rec_loss}, we employ two separate EKFs to produce these poses during training. 
\subsubsection{Pose Initialization Scheme} 

During training, each sample must be initialized with an accurate estimate of the state (and state covariance) at the first time step. 
When using supervised training, this condition is trivial to enforce: the filter is initialized with the ground truth first pose. 
Since ground truth information is not available in the self-supervised loss formulation, we initialize the training samples using the most recent pose estimate from our hybrid VIO system instead.
The pose estimates for all training sequence frames are generated at the start of every epoch, and remain fixed for each epoch.
As training progresses, the pose estimates improve, so the initialization accuracy increases.
This increase in initialization accuracy allows the training to converge further.

\begin{figure}[]
	\centering
	\begin{subfigure}[]{0.8\columnwidth}
		\includegraphics[width=\textwidth]{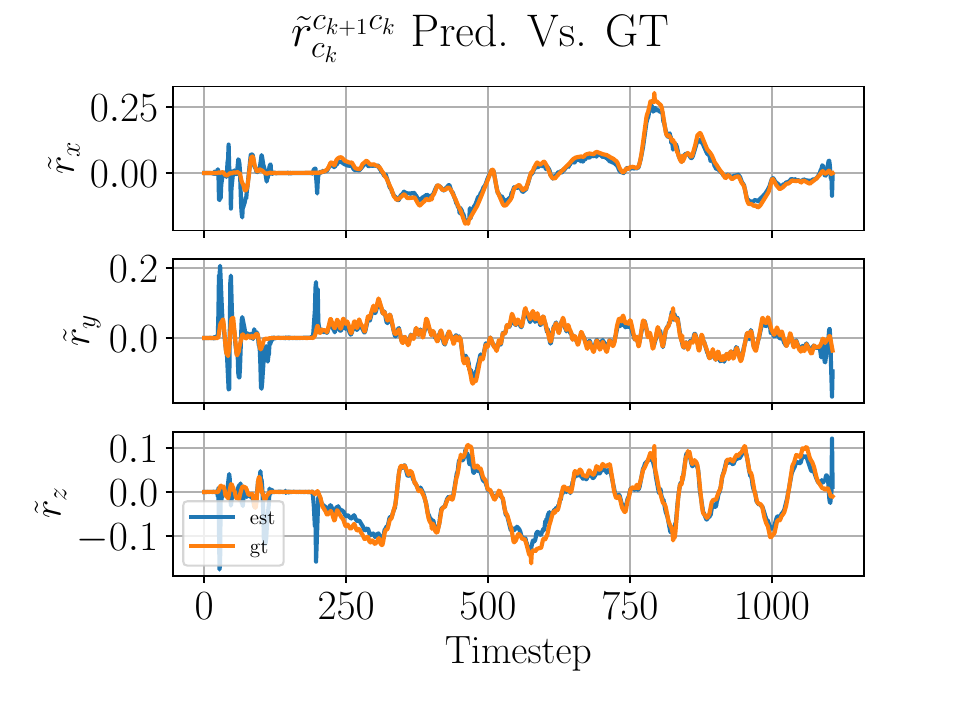}
	\end{subfigure} 
	\vspace{-0.3cm}
	\caption{The raw translation predictions for EuRoC validation sequence \texttt{MH05}. The scale of the predictions closely matches that of the ground truth, indicating that the networks are scale-aware. }
	\label{fig:trans_pred_vs_gt}
	\vspace{-5mm}
\end{figure}

To ensure a reasonable initialization for the first epoch, pose estimates are generated using a pretrained \textit{unscaled} egomotion network, which we trained using the self-supervised losses from \Cref{sec:depth_and_ego}.
To enable a metrically scaled pose initialization, a scale parameter $\lambda$ is augmented to the end of the state vector (similar to \cite{li2020thesis}).
The continuous time dynamics model for the scale is $\dot{\lambda}=0$ and error state is $\delta \dot{\lambda}=0$. The scale factor is applied to the IMU translation state, through $\lambda \Vector{r}_{r_k}^{v_{k+1} r_k}$, prior to computing $\Vector{\epsilon}_r$ within the measurement model. 
The rotation measurement is unchanged. The measurement Jacobian becomes
\begin{align}
\mathbf{H}_{k+1} = \begin{bmatrix} \ZeroMatrix_{3 \times 9} &  -\CCalT\LeftJacobianSO (-\check{\phi}) & \ZeroMatrix_{3 \times 3} & \ZeroMatrix_{3 \times 9} & \ZeroMatrix_{1 \times 3} \\[1mm]
\ZeroMatrix_{3 \times 9} & \CCalT \check{\mathbf{C}}\rCal^\wedge & -\CCalT \check{\lambda} & \ZeroMatrix_{3 \times 9} & -\CCalT \check{\mathbf{r}}\end{bmatrix},
\end{align}
where the subscripts of the state variables have been removed for brevity.
Initializing $\mathbf{x}_0$ with metric scale in the training process allows the depth and egomotion prediction scales to converge to unity.
\Cref{fig:trans_pred_vs_gt} demonstrates the scale-aware translation predictions from the trained egomotion network. 

\subsubsection{Uncertainty-Aware Measurement Model}

 To leverage the learned egomotion measurements, the iterative egomotion network \cite{wagstaff2021self} estimates a measurement covariance matrix 
\begin{equation}
\mathbf{R}_k = \begin{bmatrix} \boldsymbol{\Sigma}_{\phi_k} & \ZeroMatrix_{3 \times 3} \\ \ZeroMatrix_{3 \times 3} & \boldsymbol{\Sigma}_{r_k} \end{bmatrix},
 \end{equation}
 where $\boldsymbol{\Sigma}_{r_k}$ and $\boldsymbol{\Sigma}_{\phi_k}$ are diagonal covariance submatrices.
 Similar to \cite{li2020towards}, the dimensionality of the final layer of the network is increased from 6 to 12.
 The additional six outputs $w_i$ populate the diagonal of $\mathbf{R}_k$ through $\sigma_i^2 = \sigma^2_0 10^{\beta \tanh(w_i)}$, where $\sigma^2_0$ is a base covariance and $\beta \in \Real_{>0}$ is a control parameter.

Similar to the iterative update of the initial egomotion network predictions through multiple forward passes, the uncertainty predictions are iteratively refined.
Additional forward passes improve the alignment of the input images, which enables the network to observe discrepancies that indicate prediction errors. The network learns to inflate the predicted covariance in sequences with high discrepancy levels.

\section{Experiments \& Results}

We trained and evaluated our system on the EuRoC dataset \cite{burri2016euroc}, which includes visual-inertial data collected from an AscTec Firefly micro aerial vehicle (MAV).
The dataset consists of 11 sequences that were collected within three different environments.
The MAV underwent rapid movement in each sequence, inducing motion blur in the image stream and making this dataset challenging for odometry estimation.
The MAV was equipped with a global shutter stereo camera operating at 20 Hz and a Skybotix IMU sensor operating at 200 Hz.
Visual and inertial measurements were synchronized on-board. 
Although ground truth is available, we used these data to evaluate our self-supervised approach only.
In our experiments, the images were undistorted using the provided radial and tangental camera lens parameters and downsized to $256 \times 448$.
The left and right images from all sequences except \texttt{MH05}, \texttt{V103}, and \texttt{V203} were used during training.
The left and right images were treated as independent sequences because our method is purely monocular at training and inference time.
\begin{figure}[]
	\centering
	\includegraphics[width=0.75\columnwidth]{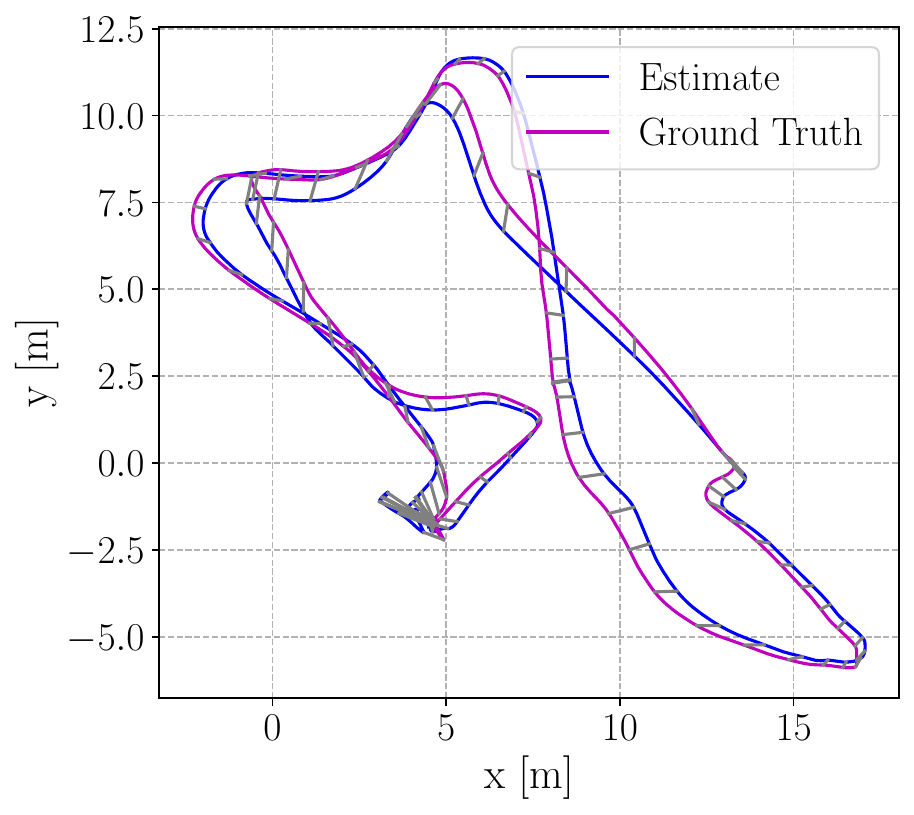}
	\vspace{-2mm}
	\caption{Top-down view of the trajectory estimate for EuRoC sequence \texttt{MH05} after \texttt{Sim3} alignment. }
	\label{fig:MH05_traj}
	\vspace{-5mm}
\end{figure}
\subsection{Training Details}

Our system was trained in PyTorch \cite{paszke:2017} using an Nvidia Quadro RTX 8000 GPU.
We used the depth and egomotion networks from \cite{wagstaff2021self}.
Initially, our model was pretrained on the ScanNet dataset, and then refined on the EuRoC training sequences.
The EuRoC IMU and image data used to train our system were downsampled by a factor of two to reduce the per-epoch training time and increase the perspective change between frames.
The training samples consisted of subsequences that were one second (10 images at 10 Hz) in length. 
Adjacent training samples had an overlap of 0.3 s.
The depth and egomotion networks were trained, in minibatches of six samples, via gradient descent (using the Adam optimizer \cite{Kingma:2014} with $\beta_1=0.9$, $\beta_2=0.999$) for 25 epochs with a learning rate of $1 \times 10^{-4}$ that was halved every seven epochs.
Five egomotion iterations (i.e., forward passes) were applied at training time and at test time.
For image augmentation during training, we randomly applied brightness, contrast, saturation, and hue transformations, in addition to random horizontal flips ($p =$ 0.5); the same augmentation was applied to every image within the same training sample.\footnote{Since the IMU data cannot be augmented in accordance with the image flipping, the egomotion predictions were altered, prior to use within the EKF, to represent the egomotion of the unflipped image.}
For training, we set $\alpha=0.15$, 
$\lambda_{photo}=1$, $\lambda_{smooth}=0.05$, and $\lambda_{consist}=0.15$.
The constants for uncertainty prediction were $\beta=4$ and $\sigma^2_0=1$.
Our IMU noise parameters were $\sigma_{\omega}= 1 \times 10^{-3}$, $\sigma_{a}=0.1$, $\sigma_{b_\omega}=1 \times 10^{-5}$, $\sigma_{b_a}=0.01$, and the covariance initialization for $\mathbf{P}_0$ used during training was $\sigma_{g_0}=0.1$, $\sigma_{v_0}=0.01$,  $\sigma_{b_{a_0}}=10$, $\sigma_{b_{\omega_0}}=0.1$, along the main diagonal. %
\begin{figure}[]
	\centering
	\begin{subfigure}[]{0.24\columnwidth}
		\includegraphics[width=\textwidth]{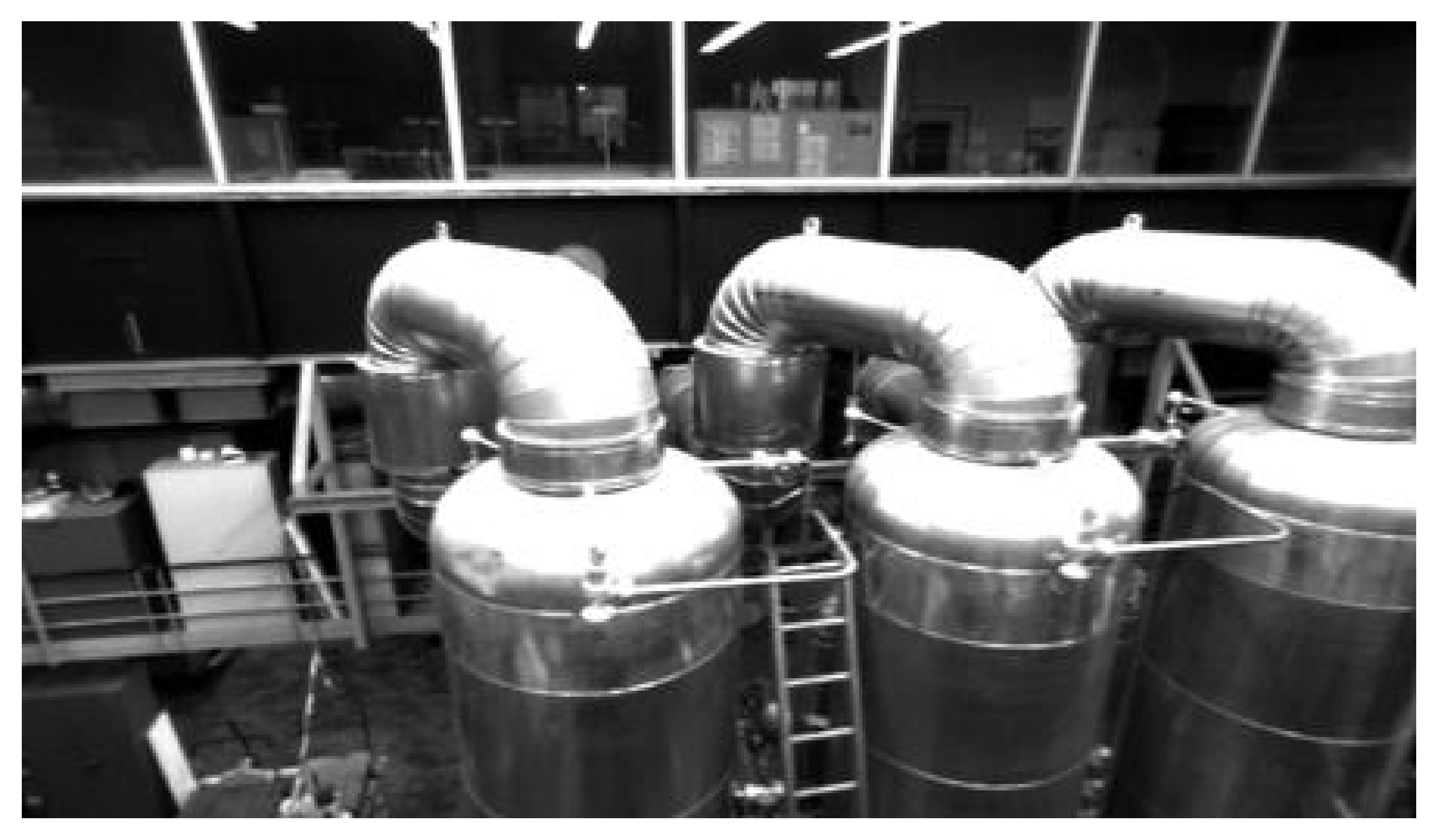}		
	\end{subfigure} 
	\begin{subfigure}[]{0.24\columnwidth}
		\includegraphics[width=\textwidth]{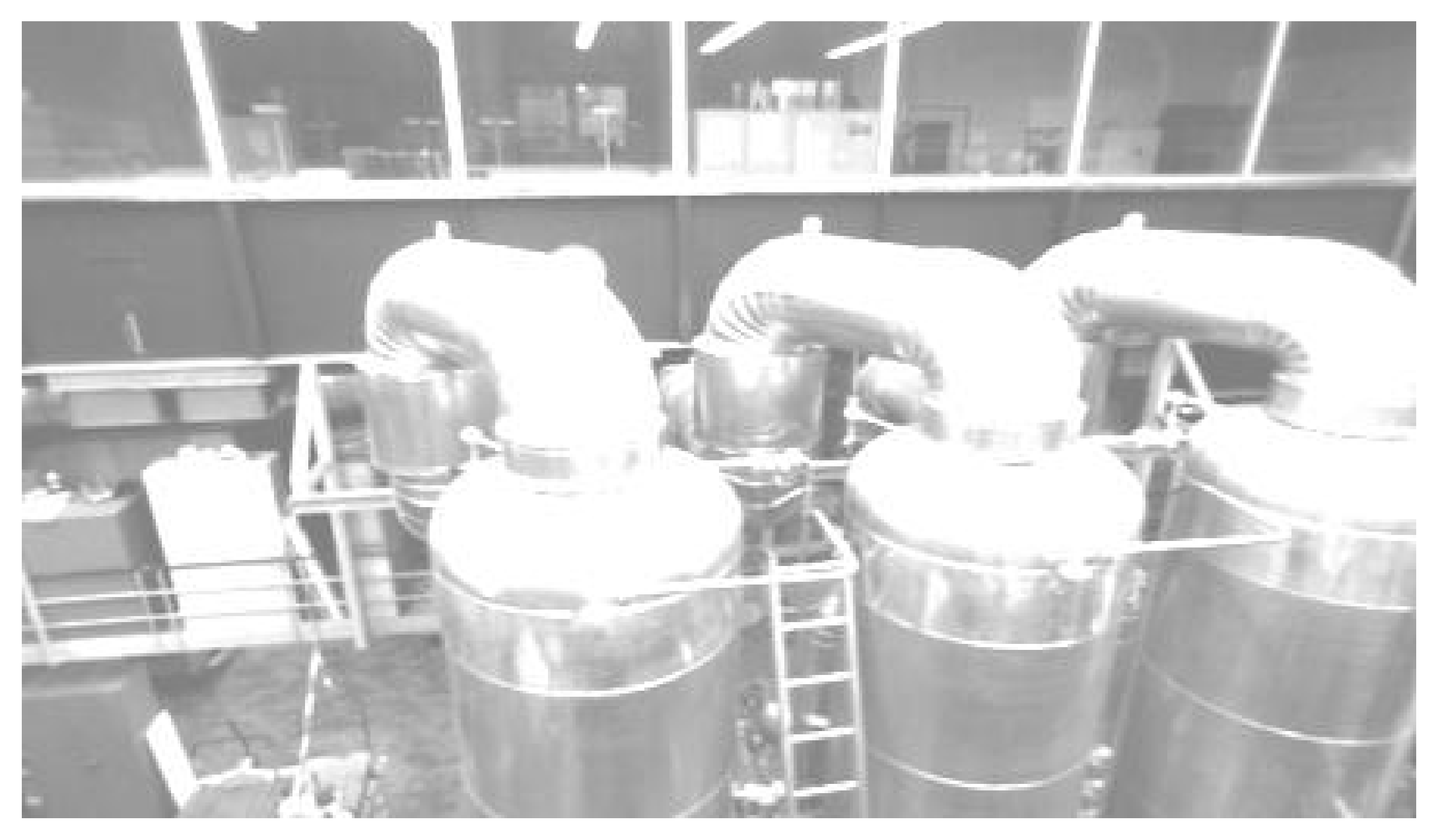}	
	\end{subfigure} 
	\begin{subfigure}[]{0.24\columnwidth}
		\includegraphics[width=\textwidth]{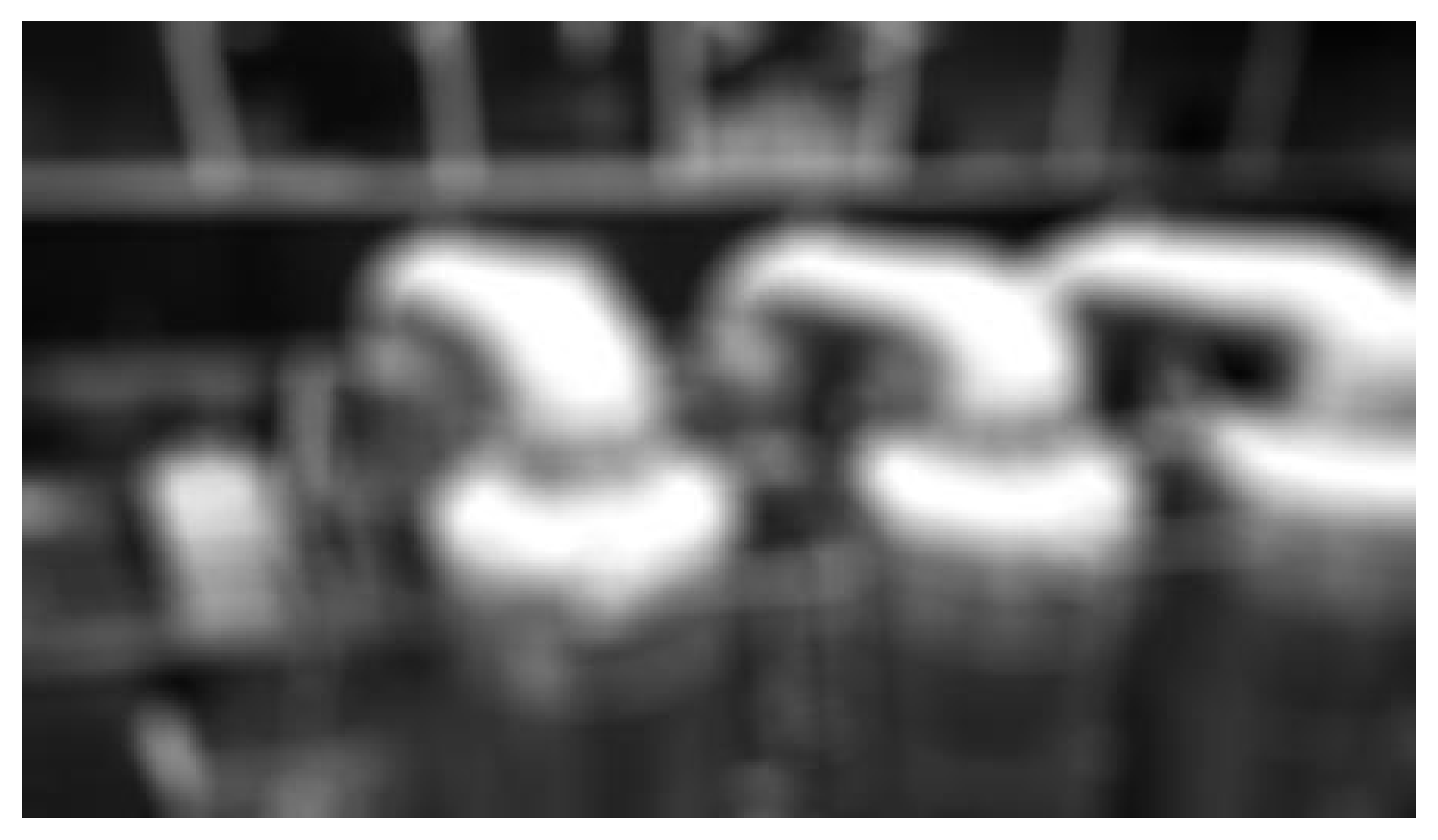}		
	\end{subfigure} 
	\begin{subfigure}[]{0.24\columnwidth}
		\includegraphics[width=\textwidth]{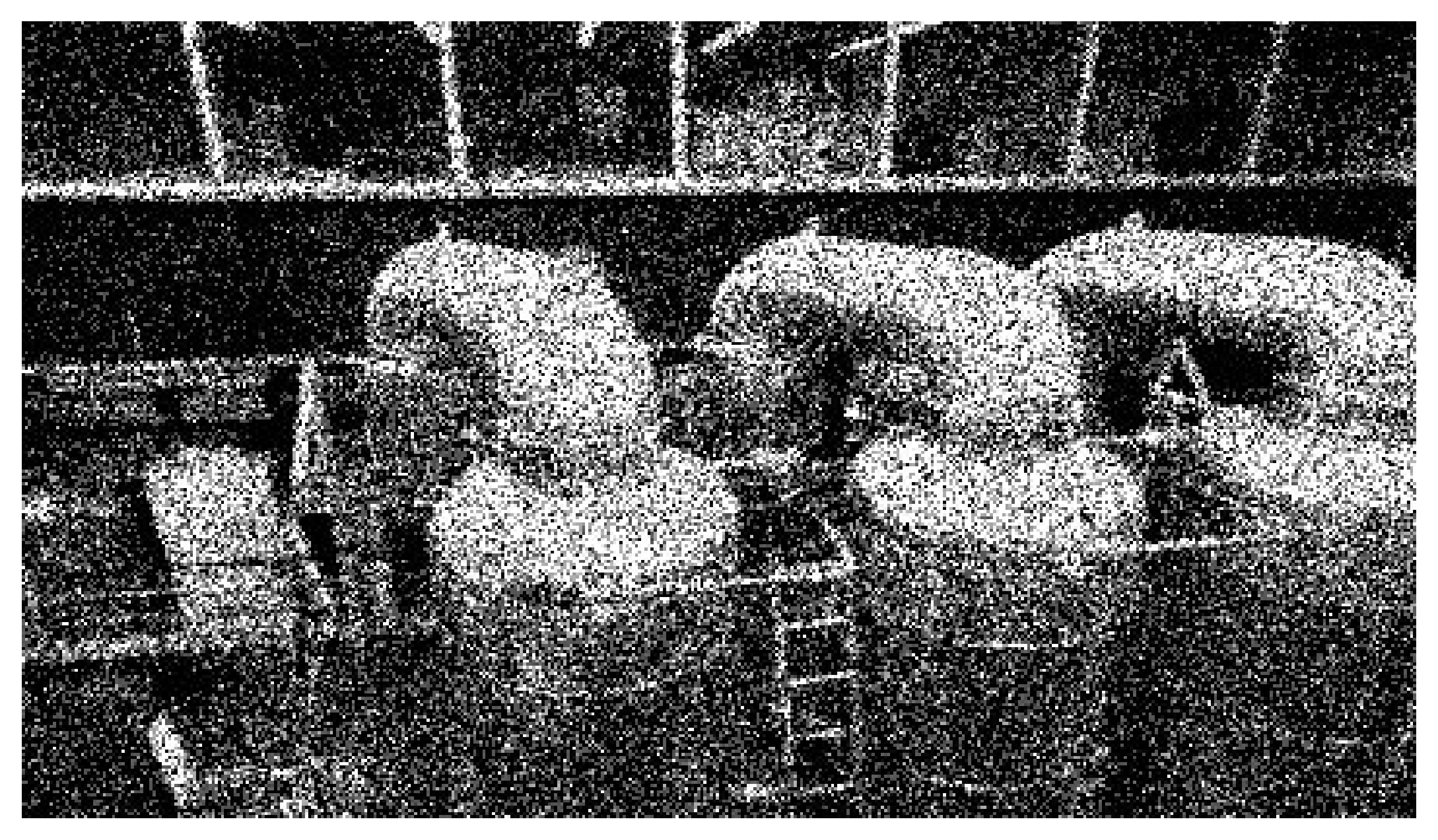}
	\end{subfigure} \\
	\begin{subfigure}[]{0.24\columnwidth}
		\includegraphics[width=\textwidth]{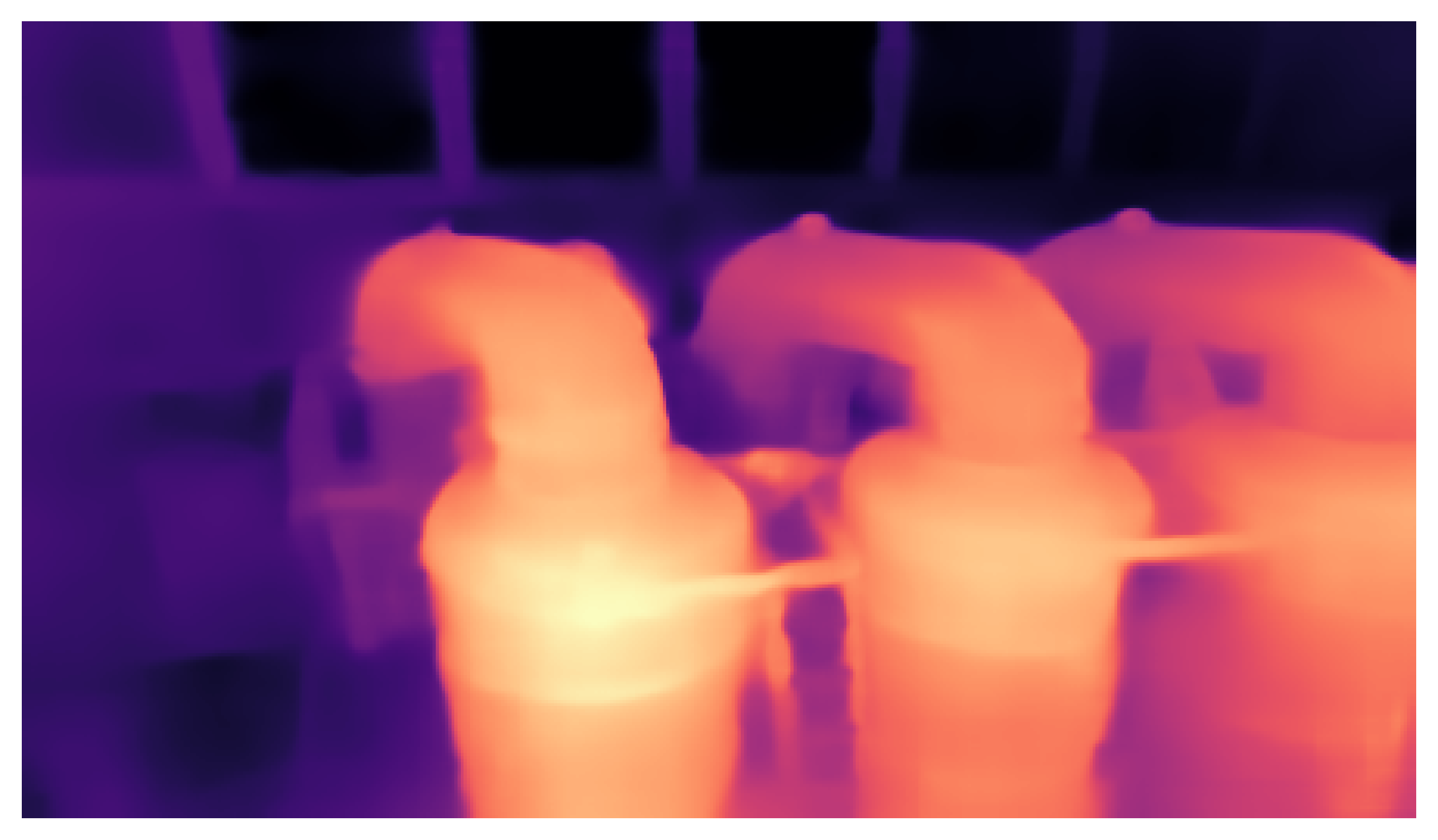}		
	\end{subfigure} 
	\begin{subfigure}[]{0.24\columnwidth}
		\includegraphics[width=\textwidth]{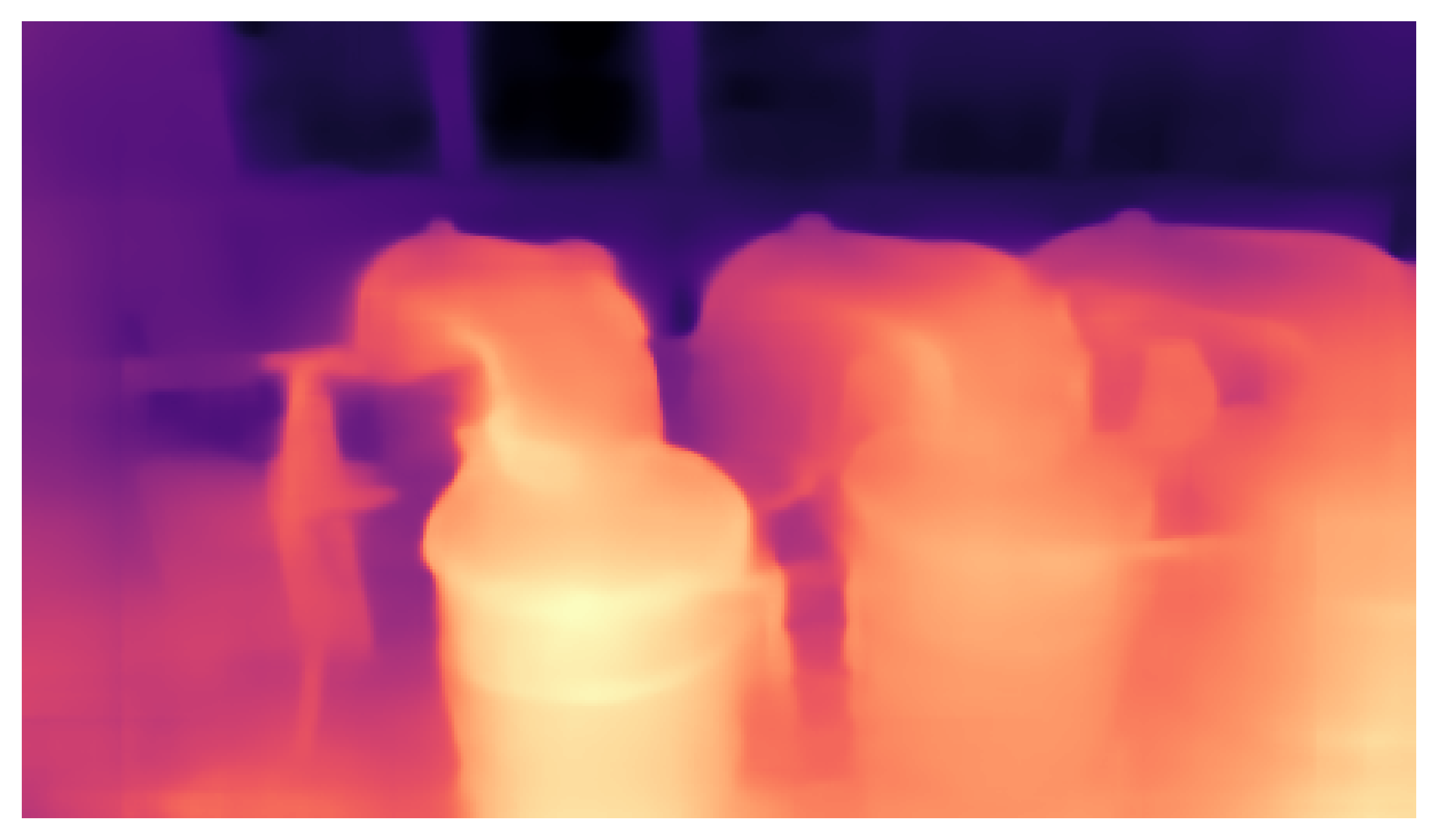}	
	\end{subfigure} 
	\begin{subfigure}[]{0.24\columnwidth}
		\includegraphics[width=\textwidth]{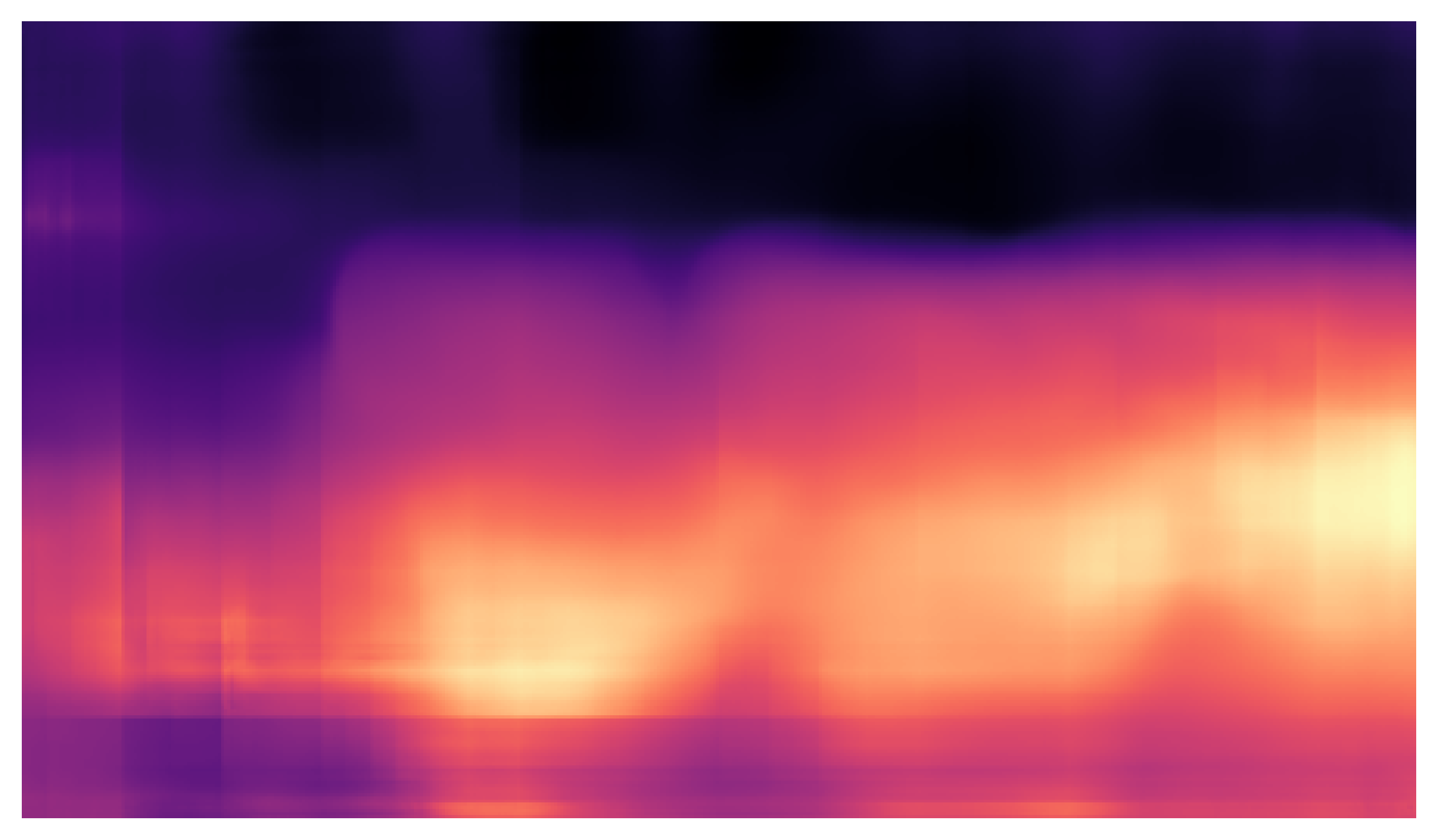}		
	\end{subfigure} 
	\begin{subfigure}[]{0.24\columnwidth}
		\includegraphics[width=\textwidth]{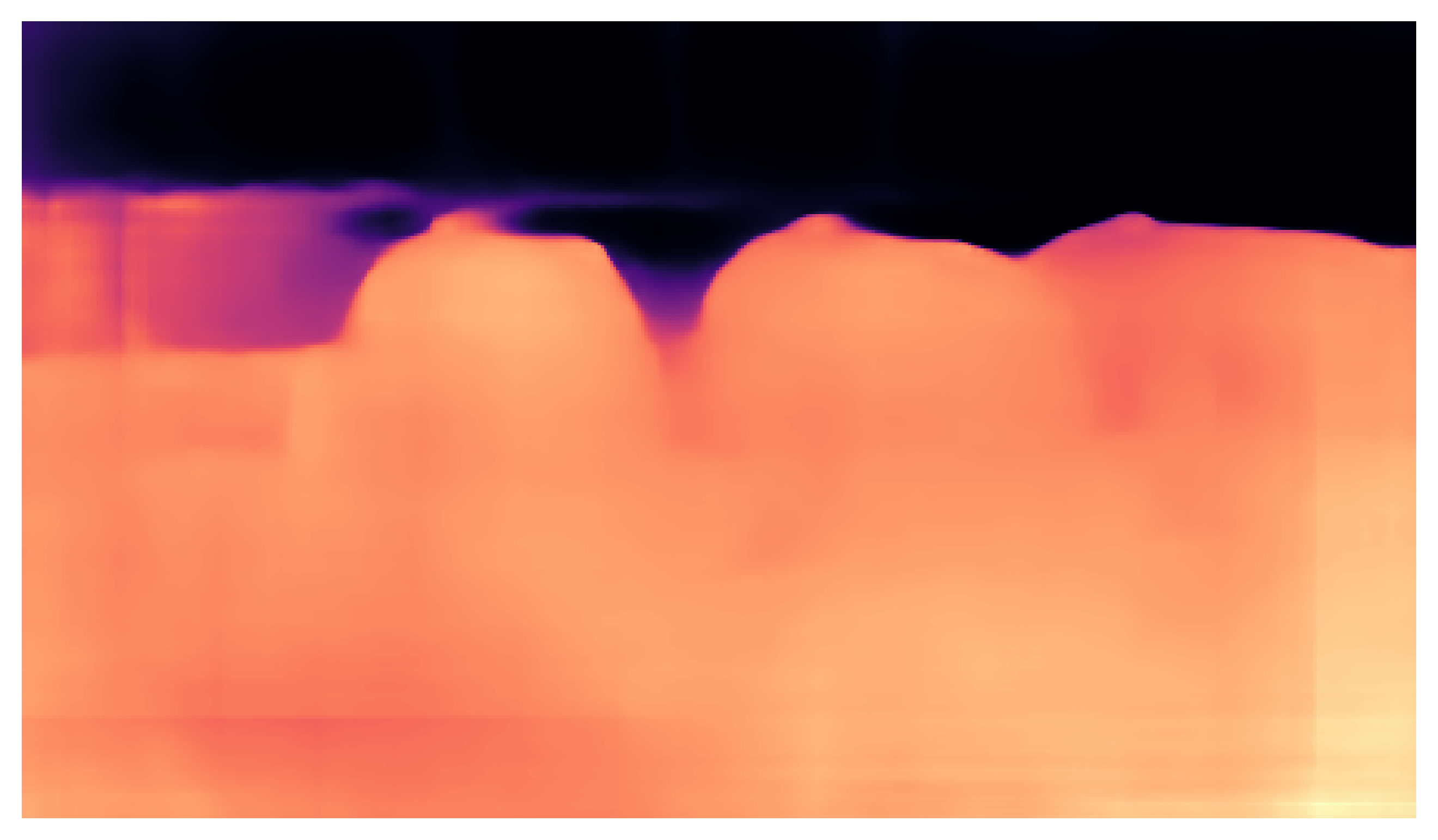} 
	\end{subfigure} \\ 
	\vspace{0.3cm}
	\begin{subfigure}[]{0.24\columnwidth}
		\includegraphics[width=\textwidth]{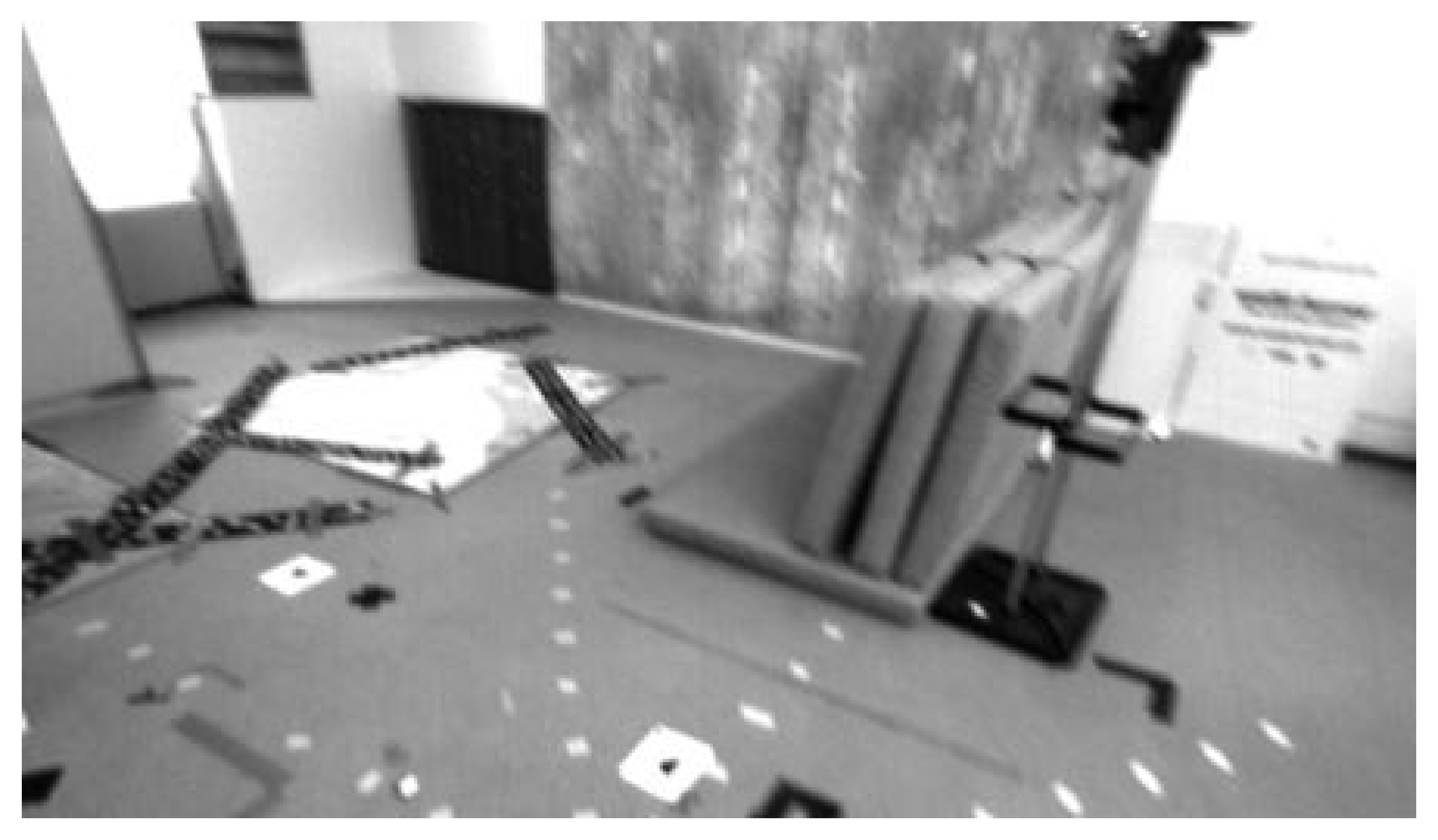}		
	\end{subfigure} 
	\begin{subfigure}[]{0.24\columnwidth}
		\includegraphics[width=\textwidth]{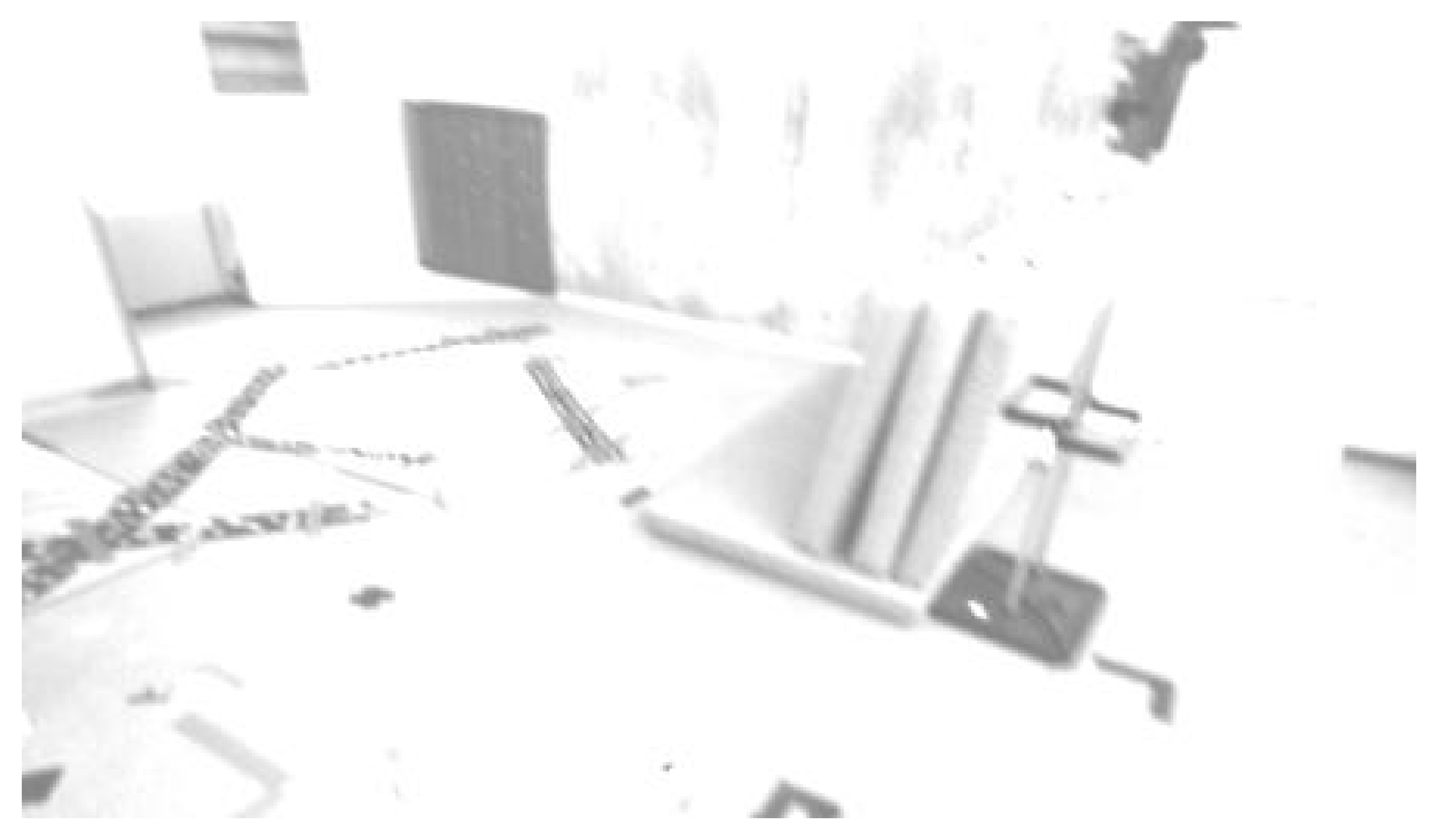}	
	\end{subfigure} 
	\begin{subfigure}[]{0.24\columnwidth}
		\includegraphics[width=\textwidth]{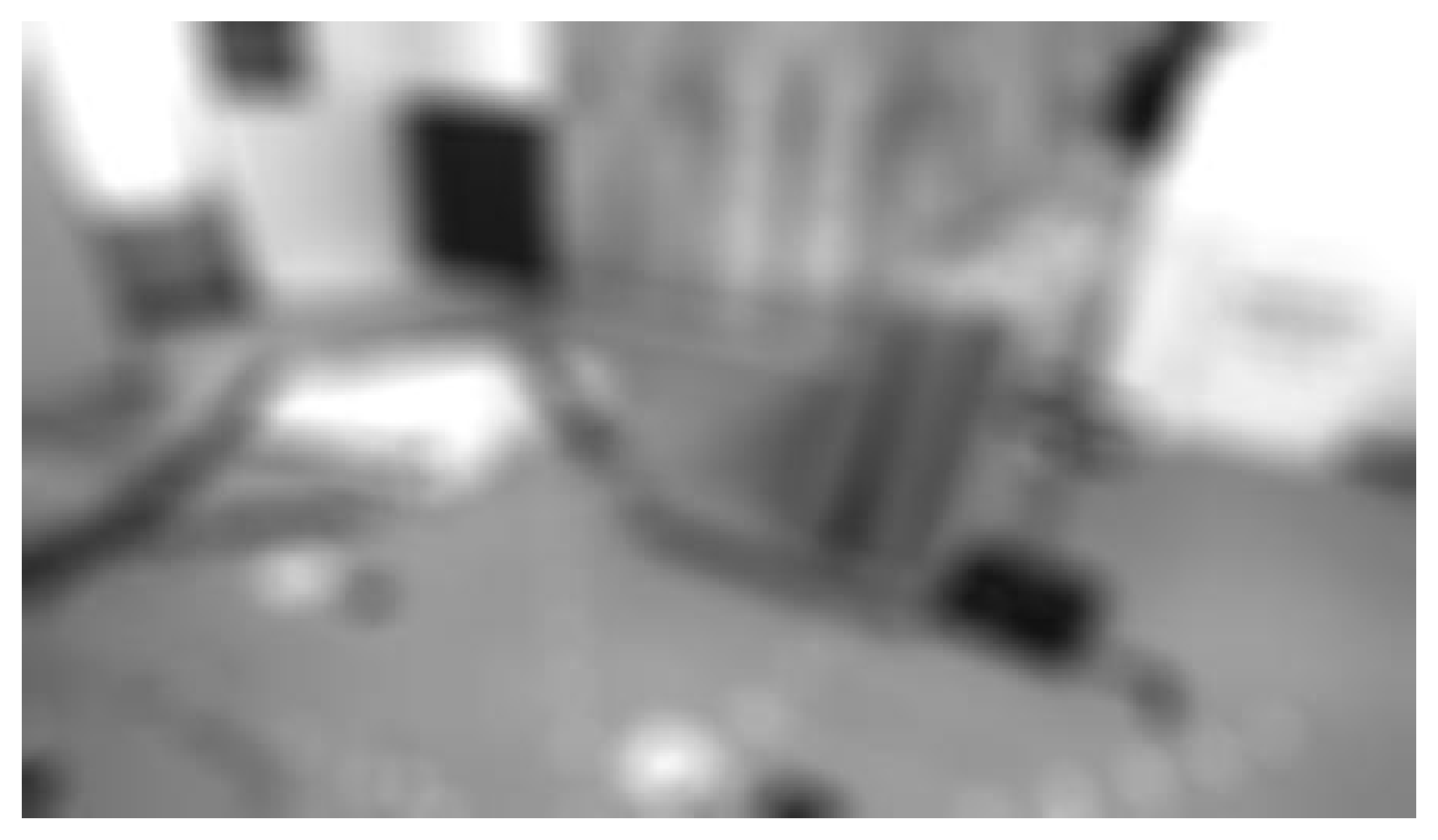}		
	\end{subfigure} 
	\begin{subfigure}[]{0.24\columnwidth}
		\includegraphics[width=\textwidth]{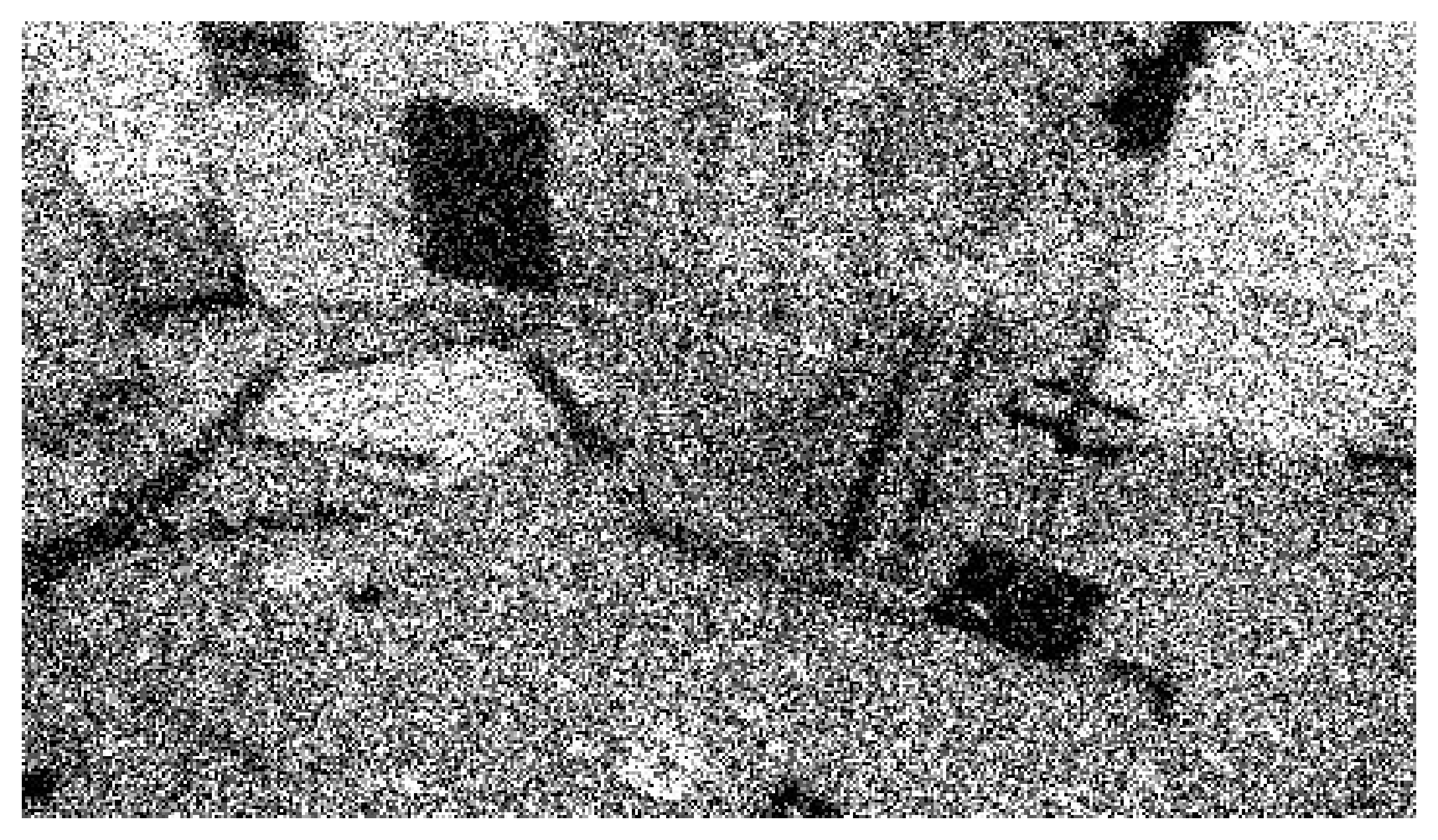}
	\end{subfigure} \\
	\begin{subfigure}[]{0.24\columnwidth}
		\includegraphics[width=\textwidth]{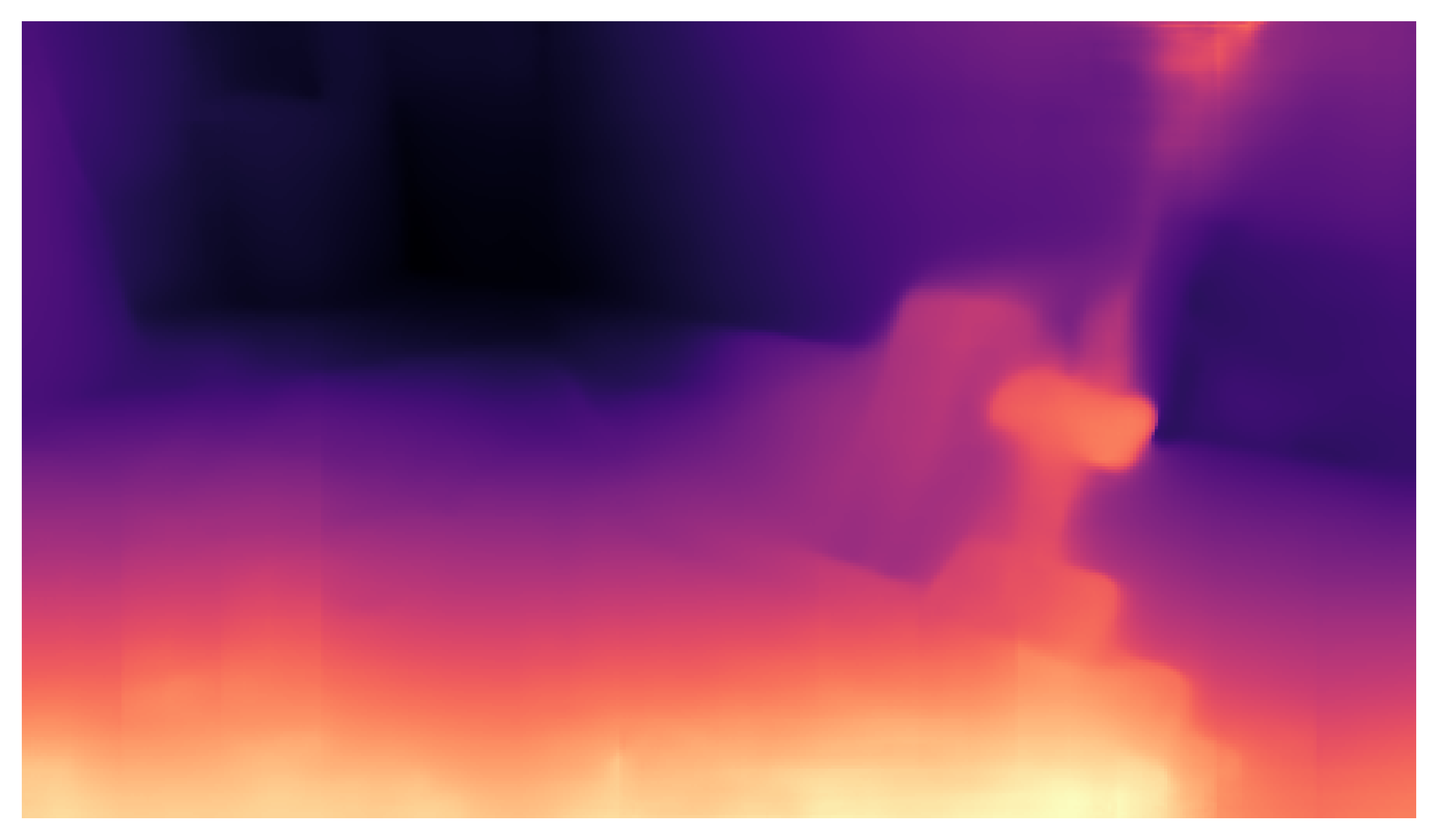}		
	\end{subfigure} 
	\begin{subfigure}[]{0.24\columnwidth}
		\includegraphics[width=\textwidth]{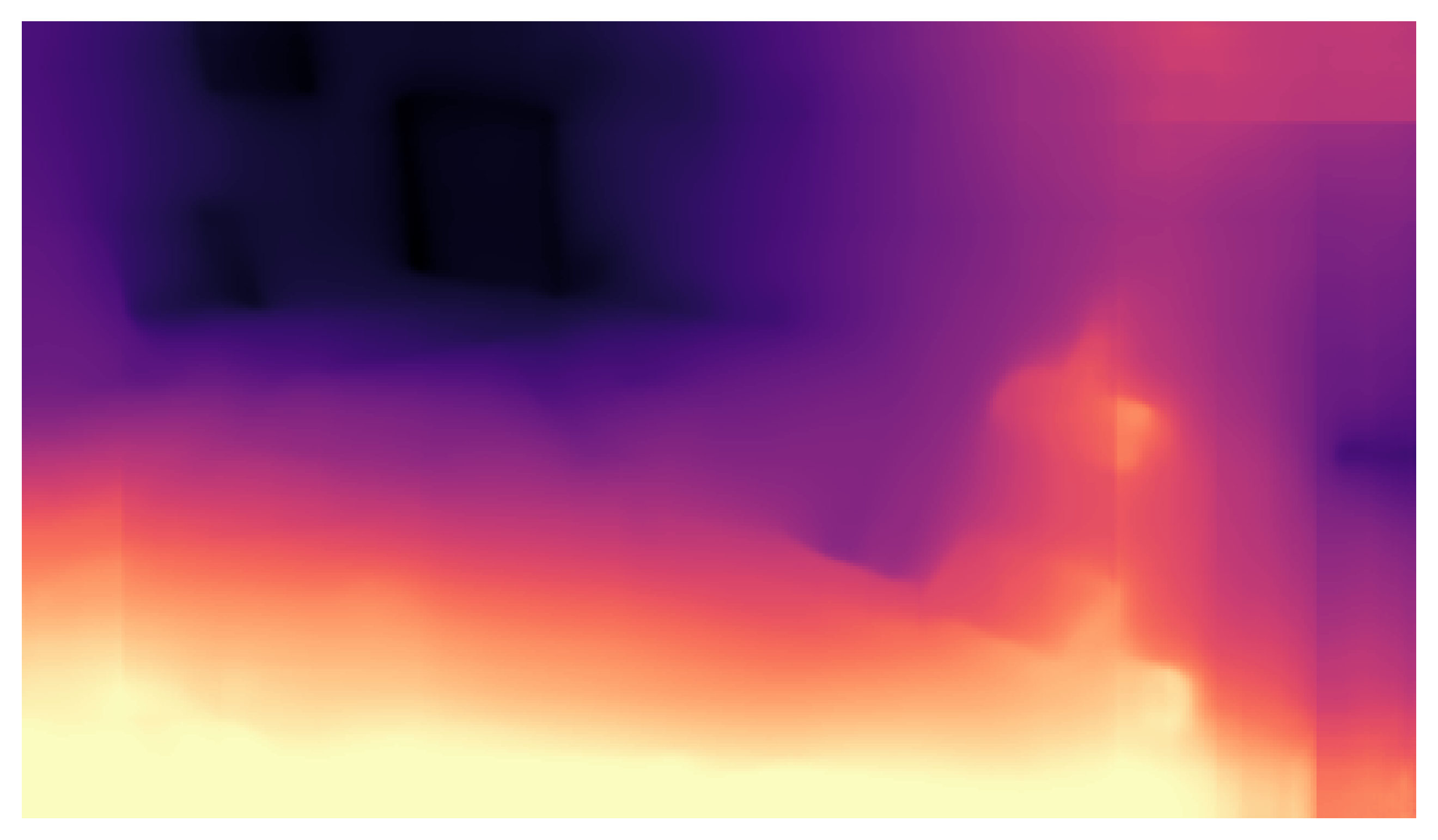}	
	\end{subfigure} 
	\begin{subfigure}[]{0.24\columnwidth}
		\includegraphics[width=\textwidth]{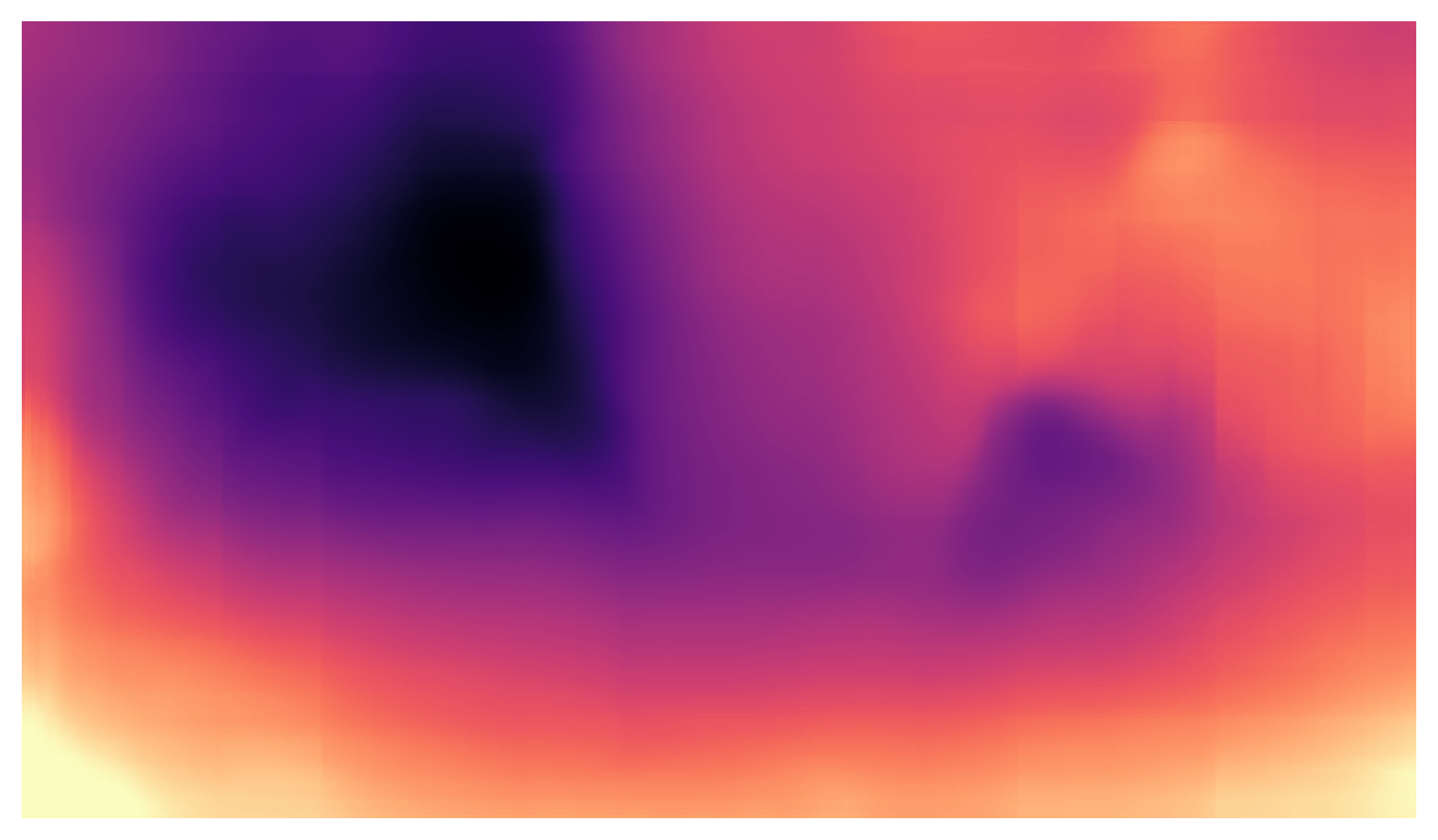}		
	\end{subfigure} 
	\begin{subfigure}[]{0.24\columnwidth}
		\includegraphics[width=\textwidth]{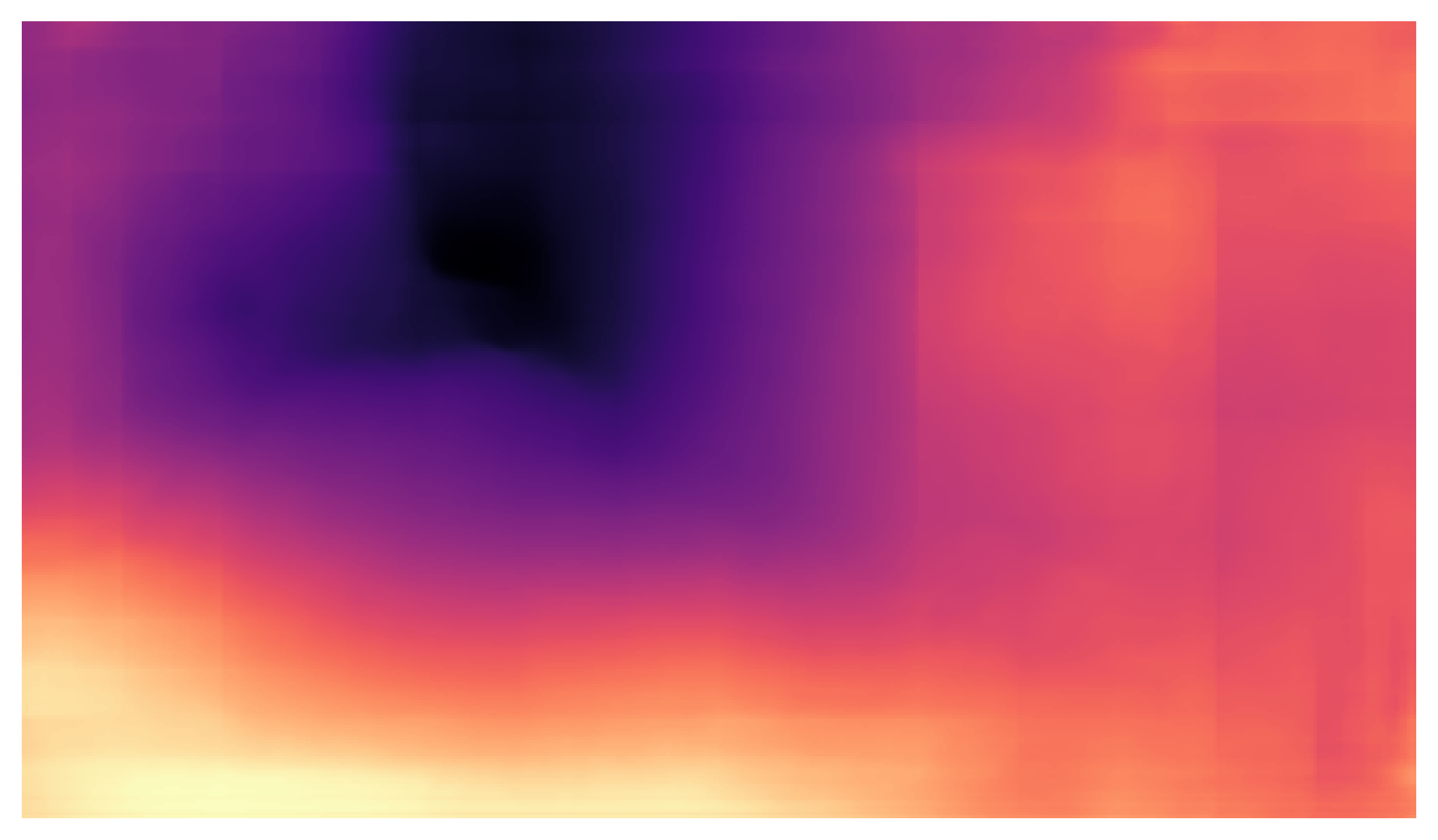} 
	\end{subfigure} \\ 
	\caption{Images from the EuRoC dataset validation sequences, and the resulting image after our applied brightness, blur, and shot noise corruptions. Despite severe image degradation, the depth network reasonably estimates the scene depth.}
	\label{fig:euroc_depth}
	\vspace{-2mm}
\end{figure}
\subsection{EuRoC Dataset Results}
\Cref{tab:euroc_results} reports the average translation RMSE, after \texttt{Sim3} alignment, for various VIO algorithms operating on the EuRoC sequences.
Similar to other self-supervised methods \cite{almalioglu2019selfvio}, we report the training sequence results.
We include comparisons with classical systems (ROVIO and VINS-Mono), a learning-based system (SelfVIO), and a hybrid system (the differentiable EKF approach from Li et al.\ \cite{li2020towards}).
\begin{table}[]
	\scriptsize
	\caption{Translation RMSE for the EuRoC sequences (using the ROVIO and VINS-Mono results reported in \cite{delmerico2018benchmark}). Our self-supervised system outperforms the supervised variant by a significant margin.}
	\label{tab:euroc_results}
	\centering
	\begin{threeparttable}
		\begin{tabular}{cccccc}
			\toprule
			Sequence & \begin{tabular}[c]{@{}c@{}} ROVIO \\ \cite{bloesch2017iterated}\end{tabular} & \begin{tabular}[c]{@{}c@{}} VINS-Mono \\ \cite{qin2018vins}\end{tabular} & \begin{tabular}[c]{@{}c@{}} SelfVIO \\ \cite{almalioglu2019selfvio} \end{tabular} & \begin{tabular}[c]{@{}c@{}} EKF-VIO \\ \cite{li2020towards} \end{tabular} & Ours \\ \midrule
			\texttt{MH01} 		& 0.21 & 0.27 & \textbf{0.19} & 1.17 & 0.51 \\
			\texttt{MH02} 		& 0.25 & \textbf{0.12} & 0.15 & 1.56 & 0.78 \\
			\texttt{MH03} 		& 0.25 & \textbf{0.13} & 0.21 & 1.89 & 0.69 \\
			\texttt{MH04} 		& 0.49 & 0.23 & \textbf{0.16} & 2.12 & 1.00 \\
			\texttt{MH05}$^\dag$ & 0.52 & 0.35 & \textbf{0.29} & 1.96 & 0.80 \\
			\texttt{V101} 		& 0.10 & \textbf{0.07} & 0.08 & 2.07 & 0.43 \\
			\texttt{V102} 		& 0.10 & 0.10 & \textbf{0.09} & 2.20 & 0.61 \\
			\texttt{V103}$^\dag$ & 0.14 & 0.13 & \textbf{0.10} & 2.83 & 0.72 \\
			\texttt{V201} 		& 0.12 & \textbf{0.08} & 0.11 & 1.49 & 0.20 \\
			\texttt{V202} 		& 0.14 &\textbf{0.08}  & \textbf{0.08 }& 2.22 & 0.81 \\
			\texttt{V203}$^\dag$ & 0.14  & 0.21 & \textbf{0.11} & --- & 0.84 \\ \bottomrule
		\end{tabular}
		\begin{tablenotes}
			\item[$\dag$] These sequences are within the held-out validation set.
		\end{tablenotes}
	\end{threeparttable}
\vspace{-5mm}
\end{table}
Notably, our system is significantly more accurate than the (supervised) hybrid approach from \cite{li2020towards}, but SelfVIO yields better performance overall.
We tentatively attribute this result to the adversarial loss applied in SelfVIO, which may lead to more accurate depth predictions.
However, SelfVIO cannot produce metrically scaled predictions and instead relies on \texttt{Sim3} alignment to recover scale.
\Cref{fig:MH05_traj} visualizes the accuracy of our system on \texttt{MH05}.

\subsection{Visual Degradation Experiments}

Next, we investigate how robust our hybrid method is to a number of realistic visual degradations.
For this experiment, our VIO approach and other VIO approaches were tested on EuRoC validation sequences but with degraded image streams.
\Cref{sec:image-corr} and \Cref{sec:frame-skip} describe the image degradations applied, while \Cref{sec:deg-results} presents our experimental results.
\subsubsection{Image Corruption}
\label{sec:image-corr}

Images were corrupted in three ways: by applying brightness transformations, by applying defocus blurring, and by adding shot noise.
The corruptions were applied using the \texttt{ImgAug} library\footnote{See \url{https://imgaug.readthedocs.io/en/latest/}}, and a severity level of five was selected. 
Only one type of corruption at a time was applied.
\Cref{fig:euroc_depth} shows various example corrupted EuRoC images.
For our experiments, we applied each corruption to all images within a window of 20 s, every 40 s (i.e., half the images were corrupted).
\subsubsection{Frame Skipping} 
\label{sec:frame-skip}

To simulate larger perspective changes, or a reduced camera/IMU frame rate, we downsampled the image and IMU data streams across the full validation sequences.
In our notation, \texttt{1:X} refers a downsample rate of \texttt{X}, where only one of \texttt{X} frames is maintained (e.g., \texttt{1:2} removes half of the frames).
We tested downsample rates of \texttt{X} $\in \{2,3,4\}$. 
\subsubsection{Degradation Experiment Results} 
\label{sec:deg-results}
\begin{figure}[]
	\centering
	\begin{subfigure}[]{0.238\textwidth}
		\includegraphics[width=\textwidth]{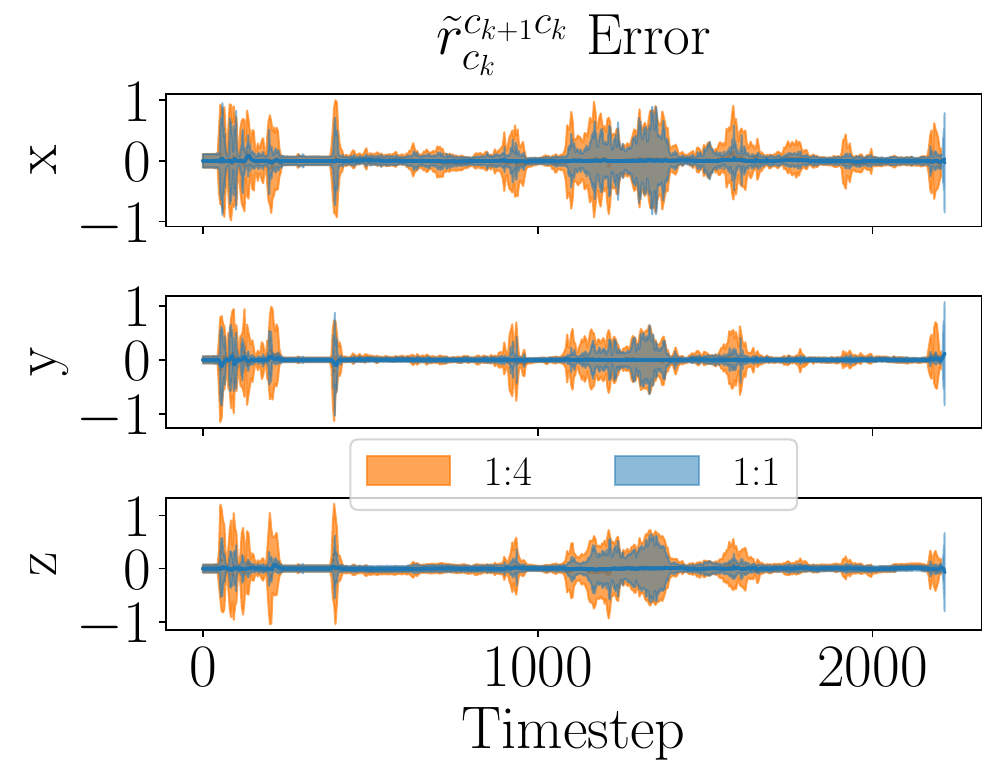}		
	\end{subfigure} 
	\begin{subfigure}[]{0.238\textwidth}
		\includegraphics[width=\textwidth]{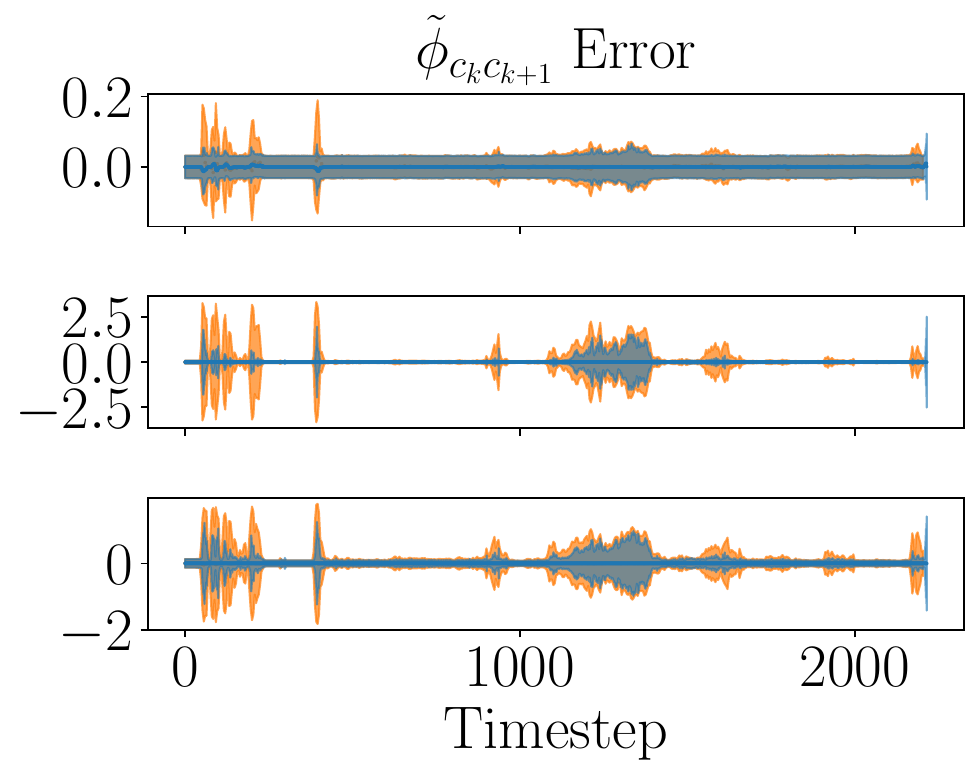} 	
	\end{subfigure}
	\caption{The learned relative egomotion measurement errors with their associated $1\sigma$ uncertainty bound for the full sequence (\texttt{1:1}), and quarter-frame sequence (\texttt{1:4}). The learned measurement covariance naturally inflates for the \texttt{1:4} sequence due to the larger perspective change between frames.}
	\label{fig:error_plots_with_covar}
	\vspace{-2mm}
\end{figure}

We evaluated our system, along with several others, on the degraded data.
We tested the classical estimators VINS-Mono \cite{qin2018vins}, ROVIO \cite{bloesch2017iterated}, and R-VIO \cite{huai2018robocentric}, and the tightly-coupled learning-based (vision-only) system \cite{wagstaff2021self} (note that SelfVIO \cite{almalioglu2019selfvio} was not publicly available).\footnote{For these comparisons, we used the open source implementations, with their default EuRoC parameters, running on Ubuntu 20.04 with ROS Noetic.}  
\Cref{tab:img_corruption_results} depicts the experimental results for sequences \texttt{MH05} and \texttt{V103}. 
We observe that degraded conditions caused the classical estimators to fail for a significant number of the trials.
During these failures, the classical estimators either lost feature tracking or diverged (resulting in 100+ metres of error).
The most robust classical system was VINS-Mono, although it could not maintain feature tracking in the frame skip experiments.
On the other hand, our system performed consistently in most trials.
Notably, the extreme \texttt{1:4} case caused little to no increase in error for our  system.
\Cref{fig:error_plots_with_covar} plots the measurement error and the predicted covariance for the \texttt{1:1} and \texttt{1:4} cases.
The covariance predictions shown in \Cref{fig:error_plots_with_covar} inflate as the camera motion becomes more extreme.
\subsection{Ablation Study}
\begin{figure}[]
	\centering
	\includegraphics[width=\columnwidth]{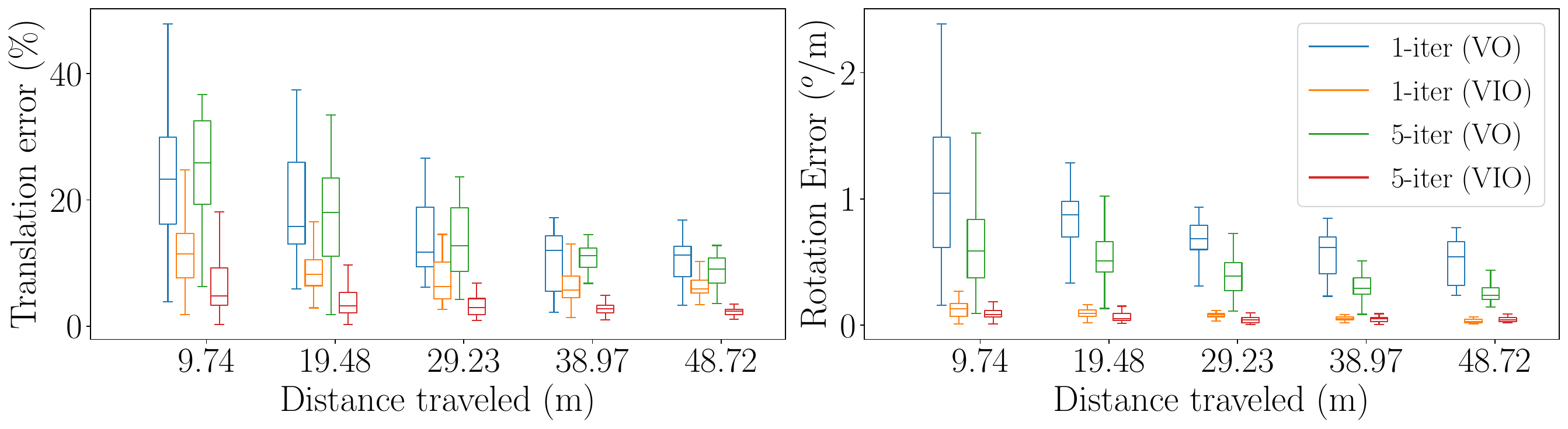}
	\caption{Ablation study showing the effect of the iterative egomotion network, and the addition of the EKF for sequence \texttt{MH05}. }
	\label{fig:iter_ekf_abl}
	\vspace{-4mm}
\end{figure}

\Cref{fig:iter_ekf_abl} shows the relative translation and rotation errors before and after the inclusion of our iterative egomotion network and our EKF components.
From \Cref{fig:iter_ekf_abl}, it is apparent that integrating the gyroscope measurements in the EKF is crucial for accurate orientation estimation.
The iterative egomotion network, when paired with the EKF, reduces the overall translation error also.

\Cref{tab:ekf_ablation} lists the mean accuracy for three versions of our system on the corrupted sequences from \Cref{tab:img_corruption_results}.
In sequence, these versions of the system are our proposed hybrid system, the standalone egomotion network (i.e., our hybrid system with the EKF removed), and the same egomotion network trained without the EKF (i.e., the baseline system from \cite{wagstaff2021self}). 
Notably, the presence of the EKF during training improves the accuracy of the raw egomotion network output. 
\begin{table*}[]
	\centering
	\scriptsize
	\vspace{3mm}
	\caption{Results for the degraded EuRoC validation sequences. We considered a failure to have occurred when 1) feature tracking was lost, or 2) the optimization diverged, resulting in rapid error growth on the order of 100+ m. Failures are indicated with an x.} 
	\label{tab:img_corruption_results}
	\begin{threeparttable}
	\begin{tabular}{lc@{\hspace{1.2\tabcolsep}}c@{\hspace{0.8\tabcolsep}}c@{\hspace{0.8\tabcolsep}}c@{\hspace{0.8\tabcolsep}}cc@{\hspace{2\tabcolsep}}c@{\hspace{0.8\tabcolsep}}c@{\hspace{0.8\tabcolsep}}c@{\hspace{0.8\tabcolsep}}c@{\hspace{0.8\tabcolsep}}c@{\hspace{0.8\tabcolsep}}}
		\toprule
		\multicolumn{1}{c}{} & \multicolumn{5}{c}{\textbf{Trans. RMSE (Sim3)}} &  & \multicolumn{5}{c}{\textbf{Rot. RMSE (Sim3)}} \\ \cmidrule{2-6} \cmidrule{8-12} 
		\multicolumn{1}{c}{\textbf{Corruption Type}} &
		Ours & Learned VO \cite{wagstaff2021self} & VINS-Mono & ROVIO & R-VIO & & Ours & Learned VO \cite{wagstaff2021self} & VINS-Mono & ROVIO & R-VIO \\ \midrule 
		\hspace{1mm}\texttt{MH05} | Nominal &  0.93 &  4.04 & \textbf{0.28}  & 1.01   & 0.46   &       & 3.57 &  11.35  & 12.95 & 4.09 & \textbf{2.36}     \\ 
		\hspace{0.75cm} Brightness &  1.11      &  4.16 & \textbf{0.29}  &  1.64  & x   &       & 6.32 &  14.56 & 15.32 & \textbf{4.38} & x     \\ 
		\hspace{0.75cm} Blur &  1.48    &  3.85 & \textbf{0.38}  & 1.20   & 0.48   &       & 5.56 &  36.06 & 13.18 & 3.47 & \textbf{1.68}     \\ 
		\hspace{0.75cm} Shot &  2.07    &  4.70 & \textbf{0.63}  & 1.54   & 0.99   &       & 12.66 &  46.77        & 14.08 & 5.55 & \textbf{4.54}     \\ [0.1cm]
		(\texttt{1:2}) | Nominal &  0.70 &  2.74 & \textbf{0.27}  & x   & x   &       & \textbf{4.77} &  9.17  & 16.28 & x & x     \\ 
		\hspace{0.75cm} Brightness &  1.09      &  2.72 & \textbf{0.34}  & x   & x   &       & \textbf{7.17} &  12.69  & 16.52 & x & x     \\ 
		\hspace{0.75cm} Blur &  1.71    &  3.92 & \textbf{0.71}  & x   & x  &       & \textbf{6.95} &  35.96 & 17.41 & x & x    \\ 
		\hspace{0.75cm} Shot &  2.43    &  4.15 &\textbf{0.67}  & x   & x   &       & \textbf{10.93} &  34.60 & 14.20 & x & x     \\ [0.1cm]
		(\texttt{1:3}) | Nominal &  0.69 &  2.18 & \textbf{0.62}  & x   & x   &       & \textbf{4.26} &  9.41  & 17.14 & x & x     \\ [0.1cm]
		(\texttt{1:4}) | Nominal &  0.67 &  1.93 & \textbf{0.41}  & x   & x   &       & \textbf{6.08} &  10.66  & 15.65 & x & x     \\ \midrule
		\texttt{V103} | Nominal &  0.43 &  0.94 & \textbf{0.15}  & 0.21   & 0.21   &       & 9.79 &  15.86 & 5.92 & \textbf{2.70} & 6.89      \\ 
		\hspace{0.75cm} Brightness &  0.53      &  0.92 & \textbf{0.17}  & 0.21   & x   &       & 9.09 &  14.17 & 6.25 & \textbf{3.88} & x      \\ 
		\hspace{0.75cm} Blur &  0.82    &  0.81 & \textbf{0.25}  & 0.26   & x   &       & 17.35 &  33.19        & 5.71 & \textbf{4.61}& x      \\ 
		\hspace{0.75cm} Shot &  0.59    &  1.33 & 0.50  & \textbf{0.23}   & x   &       & 15.12 &  124.24        & 9.50 & \textbf{3.55} & x      \\ [0.1cm]
		(\texttt{1:2}) | Nominal &  \textbf{0.72} &  \textbf{0.72} & x  & x   & x  &       & \textbf{15.07} &  15.52   & x & x & x    \\ 
		\hspace{0.75cm} Brightness &  0.86      &  \textbf{0.70} & x  & x   & x   &       & \textbf{15.56} &  16.19    & x & x & x    \\ 
		\hspace{0.75cm} Blur &  0.91    &  \textbf{0.62} & x  & x   & x   &       & \textbf{24.64} &  30.70 & x & x & x     \\ 
		\hspace{0.75cm} Shot &  \textbf{0.84}  &  1.26 & x  & x   & x   &       & \textbf{16.57} &  64.03   & x & x & x    \\ [0.1cm]
		(\texttt{1:3}) | Nominal &  0.73 &  \textbf{0.68} & x & 0.73   & x   &       & 15.77 &  17.38        & x & \textbf{8.94 }& x     \\ [0.1cm] 
		(\texttt{1:4}) | Nominal &  1.02 &  \textbf{0.88} & x  & x   & x   &       & \textbf{20.13} &  36.07       & x & x & x     \\  \midrule
		\textbf{\# Failures} &  \textbf{0}	& \textbf{0} 	& 6	& 11	& 15	& 	&\textbf{0}	& \textbf{0} 	& 6	& 11	& 15\\ 
		\textbf{ Avg.} & \textbf{1.02} & 2.16 & --- & --- & --- & & \textbf{11.37} & 29.43 & --- & --- & --- \\ \bottomrule
	\end{tabular}
\end{threeparttable}
\vspace{-3mm}
\end{table*}

\begin{table}[]
	\centering
	\scriptsize
	\caption{The mean error for three variations of our approach and the (corrupted) \Cref{tab:img_corruption_results} sequences. For the vision-only approaches, the network trained with our hybrid system is more accurate than the same network trained without the EKF.}
	\label{tab:ekf_ablation}
	\begin{threeparttable}
	\begin{tabular}{c@{\hspace{0.9\tabcolsep}}c@{\hspace{0.9\tabcolsep}}cc@{\hspace{0.8\tabcolsep}}c@{\hspace{0.8\tabcolsep}}c@{\hspace{0.8\tabcolsep}}c}
		\toprule
		\multicolumn{3}{c}{\textbf{Trans. RMSE}} &  & \multicolumn{3}{c}{\textbf{Rot. RMSE}} \\ \cmidrule{1-3} \cmidrule{5-7} 
		Ours & \begin{tabular}[c]{@{}c@{}}Ours\\ (w/o EKF)\end{tabular} & Learned VO \cite{wagstaff2021self} &  & Ours & \begin{tabular}[c]{@{}c@{}}Ours\\ (w/o EKF)\end{tabular} & Learned VO \cite{wagstaff2021self}  \\ \midrule
		\textbf{1.02}	& 1.52 & 2.16 &  & \textbf{11.37} & 17.09 & 29.43 \\ \bottomrule
	\end{tabular}
\end{threeparttable}
\vspace{-6mm }
\end{table}
\section{Conclusion}
We have demonstrated how a self-supervised, hybrid VIO system is able to effectively maintain an accurate egomotion estimate even when operating under significantly degraded conditions.
Our system, which can be trained end-to-end with a self-supervised reconstruction loss, is able to learn a heteroscedastic measurement covariance model that downweights unreliable visual measurements.
Combined with an IMU-based process model in a differentiable EKF, our principled sensor fusion scheme increases the overall estimation accuracy and allows for consistent performance.
A noteworthy attribute of our system is its ability to recover the metric scene scale. 

As future work, we intend to improve our depth predictions to boost the accuracy of the iterative egomotion network.
We plan to do so by incorporating the discriminative loss from \cite{almalioglu2019selfvio} and by investigating how raw depth predictions can be refined within a differentiable filter.
\vspace{-0.15cm}
\section*{Acknowledgments}
We are grateful to NVIDIA Corporation for providing the Quadro RTX 8000 GPU used for this research.
%
\vspace{-0.1cm}
\bibliographystyle{IEEEcaps}
\bibliography{references.bib}

\end{document}